\documentclass{article}

\usepackage[preprint]{colm2025_conference}

\usepackage{xcolor}

\usepackage[utf8]{inputenc} 
\usepackage[T1]{fontenc}    
\usepackage{hyperref}       
\usepackage{url}            
\usepackage{booktabs}       
\usepackage{amsfonts}       
\usepackage{nicefrac}       
\usepackage{microtype}      

\usepackage{booktabs,tabularx, colortbl} 

\usepackage{graphicx}
\usepackage{multirow}
\usepackage{enumitem}

\usepackage{float}

\usepackage{amsfonts} 
\usepackage{bm}
\usepackage{mathtools}
\usepackage{amsmath}

\usepackage{amsmath,amsfonts,bm}

\def\eqref#1{equation~\ref{#1}}

\def\1{\bm{1}}

\def\mA{{\bm{A}}}

\def\mK{{\bm{K}}}

\def\mQ{{\bm{Q}}}

\def\mV{{\bm{V}}}

\DeclareMathAlphabet{\mathsfit}{\encodingdefault}{\sfdefault}{m}{sl}
\SetMathAlphabet{\mathsfit}{bold}{\encodingdefault}{\sfdefault}{bx}{n}

\def\gD{{\mathcal{D}}}

\newcommand{\R}{\mathbb{R}}

\usepackage{xspace}

\newcommand{\method}[0]{PyramidKV\xspace}

\usepackage{cleveref}
\newcommand{\Scref}[1]{\S\ref{#1}}

\title{PyramidKV: Dynamic KV Cache Compression based on Pyramidal Information Funneling}

\author{%
Zefan Cai$^{1}$,
Yichi Zhang$^{2}$,
Bofei Gao$^{2}$,
Yuliang Liu$^{3}$, 
Yucheng Li$^{4}$, 
Tianyu Liu$^{5}$, \\
\textbf{Keming Lu}$^5$, 
\textbf{Wayne Xiong}$^{7}$,
\textbf{Yue Dong}$^{6}$,
\textbf{Junjie Hu}$^{1}$,
\textbf{Wen Xiao}$^{7}$,
\\
$^1$University of Wisconsin - Madison 
$^2$Peking University 
$^3$Nanjing University \\
$^3$University of Surrey
$^5$Qwen
$^6$University of California - Riverside 
$^7$Microsoft \\
  \texttt{zefncai@gmail.com} \\
\url{https://github.com/Zefan-Cai/PyramidKV} \\
}

\begin{document}

\maketitle
\begin{abstract}

In this study, we investigate whether attention-based information flow inside large language models (LLMs) is aggregated through noticeable patterns for long context processing.  Our observations reveal that LLMs aggregate information through 
\textbf{Pyramidal} Information Funneling where attention is scattering widely in lower layers, progressively consolidating within specific contexts, and ultimately focusing on critical tokens (a.k.a massive activation or attention sink) in higher layers. Motivated by these insights, we developed \method, a novel and effective KV cache compression method. This approach dynamically adjusts the KV cache size across different layers, allocating more cache in lower layers and less in higher ones, diverging from traditional methods that maintain a uniform KV cache size.
Our experimental evaluations, utilizing the LongBench benchmark, show that \method matches the performance of models with a full KV cache while retaining only 12\% of the KV cache, thus significantly reducing memory usage. In scenarios emphasizing memory efficiency, where only 0.7\% of the KV cache is maintained, \method surpasses other KV cache compression techniques, achieving up to a 20.5 absolute accuracy improvement on TREC dataset. In the Needle-in-a-Haystack experiment, \method outperforms competing methods in maintaining long-context comprehension in LLMs; notably, retaining just 128 KV cache entries enables the LLAMA-3-70B model to achieve 100.0 Acc. performance.
\end{abstract}

\section{Introduction}
\label{section:introduction}

Large language models (LLMs)~\citep{achiam2023gpt,touvron2023llama,touvron2023llama2,jiang2023mistral} are integral to various natural language processing applications, including dialogue systems~\citep{vicuna2023}, document summarization~\citep{fabbri-etal-2019-multi}, and code completion~\citep{roziere2023code}. These models have recently been scaled up to handle long contexts~\citep{fu2024data, ding2024longrope, zhu2023pose, chen2023longlora}, with GPT-4 processing up to 128K tokens and Gemini-pro-1.5 handling 1M tokens. However, scaling LLMs to extremely long contexts naturally leads to a significant delay due to the quadratic computation of attention over long contexts. A common solution to mitigate such inference delays involves caching the key and value states (KV) of previous tokens~\citep{waddington2013kv}, with the trade-off of requiring extensive GPU memory storage. For instance, maintaining a KV cache for 100K tokens in LLaMA-2 7B requires over 50GB of memory, while a 2K context requires less than 1GB of memory~\citep{wu2024retrieval}.

\begin{figure}
    \centering
    \includegraphics[width=\columnwidth]{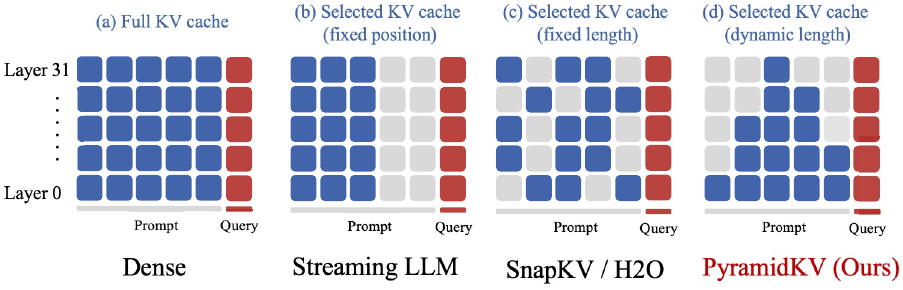}
    \vspace{-0.5cm}
    \caption{
    Illustration of~\method compared with existing KV cache compression methods.
    (a) Full KV has all tokens stored in the KV cache in each layer; cache size increases as the input length increases.
    (b) StreamingLLM~\citep{xiao2023efficient} only keeps few initial tokens with a fixed cache size in each layer.
    (c) SnapKV~\citep{li2024snapkv} and H2O~\citep{zhang2024h2o} keep a fixed cache size across Transformer layers, and their selection is based on the attention score.
    (d)~\method maintains pyramid-like cache sizes, allocating more cache budget to lower layers and less to higher layers. This approach to KV cache selection better aligns with the increasing attention sparsity observed in multi-layer Transformers (\Scref{section:observation}).
    }
    \vspace{-0.5cm}
    \label{figure:comparison}
\end{figure}

To tackle these memory constraints, recent studies have explored the optimization of KV caching, including approaches such as low-rank decomposition of the KV cache~\citep{dong2024get} or pruning non-essential KV cache~\citep{zhang2024h2o, li2024snapkv, ge2023model}.  Notably, it has been shown that maintaining merely 20\% of the KV cache can preserve a substantial level of performance~\citep{zhang2024h2o}. Moreover, extreme compression of the KV cache for tasks of longer contexts (e.g., retrieval augmented generation or RAG for short) can drastically improve efficiency and further reduce resource use. However, questions about the universal applicability of these strategies across all layers of an LLM remain open. (1) \textit{Are these KV cache strategies applicable to all layers? }(2) \textit{Is it computationally efficient to use the same KV cache size across layers as previous studies have done?} These considerations suggest a need for an in-depth, more nuanced understanding of KV cache optimization in LLMs. 

To examine these questions, we aim to systematically investigate the design principles of the KV cache compression across different layers, specifically tailored to the behaviors of the attention mechanism. We first investigate how information flow is aggregated via attention mechanisms across different layers in multi-document question answering (QA), a classic task involving long contexts. Our analysis identifies a notable transition of attention distribution from a broad coverage of global contexts to a narrow focus of local tokens over layers in LLMs. This pattern suggests an aggregated information flow where information is initially gathered broadly and subsequently narrowed down to key tokens, epitomizing the massive attention phenomenon. Our findings provide unique insights beyond the previously documented ``massive activation''~\citep{sun2024massive} that very few activations exhibit significantly larger values than others when calculating multi-head attention in LLMs and ``attention sink''~\citep{xiao2023efficient} that keeping the KV of initial tokens will largely recover the performance of window attention.

Building on these insights on how information flows are aggregated through a pyramid pattern, we design a novel and effective KV cache pruning approach that mirrors the geometric shape, named PyramidKV. As shown in \autoref{figure:comparison}, unlike the fixed-and-same length KV cache pruning common in prior works~\citep{zhang2024h2o, ge2023model, li2024snapkv}, PyramidKV allocates more KV cache to the lower layers where information is more dispersed and each KV holds less information while reducing the KV cache in higher layers where information becomes concentrated in fewer key tokens. To the best of our knowledge, PyramidKV is the first KV cache compression method with varied cache retention across layers, tailoring cache amounts to the informational needs of each layer.

\begin{figure}
    \centering
    \includegraphics[width=\columnwidth]{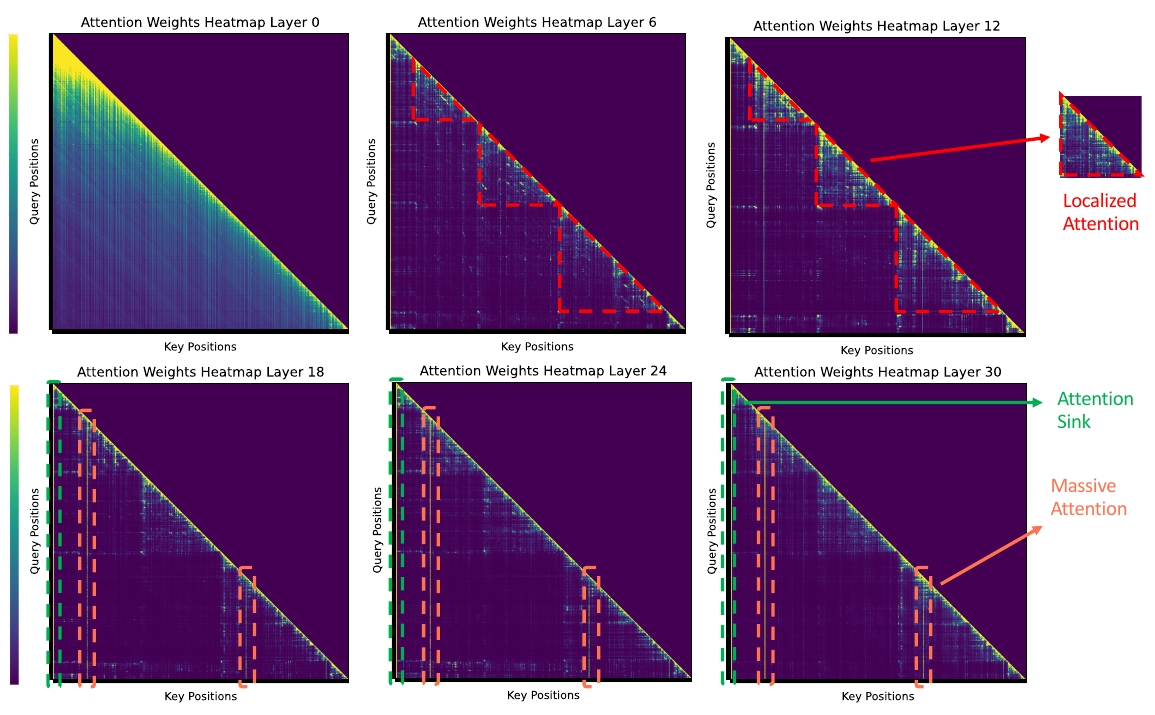}
    \vspace{-0.5cm}
    \caption{
  Attention patterns of retrieval-augmented generation across layers in LlaMa~\citep{touvron2023llama, touvron2023llama2} reveal that in the lower layers, the model exhibits a broad-spectrum mode of attention, distributing attention scores uniformly across all content. In the middle layers, attention becomes more localized within each document, indicating refined information aggregation (dotted red triangular shapes in layers 6 and 10). This culminates in the upper layers, where ``massive attention'' focuses on a few key tokens (concentrated attention bars after layer 18), efficiently extracting essential information for answers.
    }
    \vspace{-0.5cm}
    \label{figure:rag_flow}
\end{figure}

We conducted comprehensive experiments on LongBench~\citep{bai2023longbench} using 17 datasets across various tasks and domains with three backbone models (LLaMa-3-8B-Instruct, LLaMa-3-70B-Instruct and Mistral-7B~\citep{jiang2023mistral}). The results show that \method preserves performance using just 12.0\% of the KV cache (KV Cache size = 2048) on the LongBench benchmark and significantly outperforms other methods in extreme conditions, retaining only 0.7\% of the KV cache. Moreover, \method outperforms baseline models (H2O~\citep{zhang2024h2o}, SnapKV~\citep{li2024snapkv}, StreamingLLM~\citep{xiao2023efficient}) across all tested cache sizes (64, 96, 128, 256), with its advantages most pronounced at smaller cache sizes. In the Needle In A Haystack experiment, \method effectively maintains the long-context comprehension in LLMs, outperforming than competing methods. Remarkably, with \method, retaining only 128 KV cache entries allows the LLaMa-3-70B-Instruct model to achieve 100.0 Acc. performance, matching the performance of a full KV cache.
\section{Related Work}
\label{section:related_work}

There has been a growing interest in addressing LLMs' memory constraints on processing long context inputs. FastGen~\citep{ge2023model} introduces an adaptive KV cache management strategy that optimizes memory use by tailoring retention tactics to the specific nature of attention heads. 
SnapKV~\citep{li2024snapkv} improves efficiency by compressing KV caches via selecting/clustering significant KV positions based on their attention scores.  Heavy Hitter Oracle (H2O)~\citep{zhang2024h2o} implements a dynamic eviction policy that effectively balances the retention of recent and historically significant tokens, optimizing memory usage while preserving essential information.  StreamingLLM~\citep{xiao2023efficient} enables LLMs trained on finite attention windows to handle infinite sequence lengths without fine-tuning, thus expanding the models' applicability to broader contexts.
\section{Pyramidal Information Funneling}
\label{section:observation}

To systematically understand the attention mechanism over layers in LLMs for long-context inputs, we conduct a fine-grained study focusing on the multi-document question answering (QA) task. The model is presented with multiple interrelated documents and prompted to generate an answer for the given query. The main target is to investigate how the model aggregates dispersed information within these retrieved documents for accurate responses.


In particular, we focus on our analysis of the LLaMa~\citep{touvron2023llama, touvron2023llama2} and visualize the distribution and behavior of attention scores over layers.  
To assess the distinct behaviors of each multi-head self-attention layer, we compute the average attention from all heads within each layer. \autoref{figure:rag_flow} shows the attention patterns of one QA example over six different layers (i.e., 0, 6, 12, 18, 24, and 30).

We identify an approximately uniform distribution of attention scores from the lower layers (e.g., the 0th layer). This suggests that the model operates in a broad-spectrum mode at the lower layers, aggregating information globally from all available content without prioritizing its attention on specific input segments. Notably, a distinct transition to a more localized attention pattern within each document emerges, as the model progresses to encode information at the middle layers (6th to 18th layers). In this phase, attention is predominantly directed towards tokens within the same document, suggesting a more refined aggregation of information within individual contexts.

This trend continues and intensifies in the upper layers (from the 24th to the 30th layer), where we observed the emergence of `massive attention' phenomena. In these layers, the attention mechanism concentrates overwhelmingly on a few key tokens. This pattern of attention allocation, where extremely high attention scores are registered, signifies that the model has aggregated the essential information into these focal tokens. Such behavior underscores a sophisticated mechanism by which LLMs manage and streamline complex and voluminous information, culminating in the efficient extraction of the most pertinent data points necessary for generating accurate answers.

\section{PyramidKV}
\label{section:method} 
\subsection{Preliminaries and Problem Formulation}

In an autoregressive transformer LLM, the generation of the 
$i$-th token requires that the attention module computes the query, key, and value vectors for all previous $i-1$ tokens. To speed up inference process and avoid duplicate computations, the key and value matrices are typically stored in the GPU memory. While the KV cache enhances inference speed and reduces redundant computations, it can consume significant memory when dealing with long input contexts. To optimize memory usage, a strategy called KV cache compression is proposed~\citep{zhang2024h2o, xiao2023efficient, li2024snapkv}, which involves retaining only a minimal amount of KV cache while preserving as much information as possible.



In a LLM with $m$ transformer layers, we denote the key and value matrices in the $l$-th attention layer respectively as $\mK^l, \mV^l \in \R^{n\times d}, \forall l\in [0, m-1]$ when encoding a sequence of $n$ tokens. The goal of KV cache compression is to seek two sub-matrices $\mK^l_s, \mV^l_s \in \R^{k^l \times d}$ from  the full matrices $\mK^l$ and $\mV^l$, given a cache budget $k^l<n$ for each layer $l\in [0, m-1]$ while maximizing performance preservation. A LLM with KV cache compression only uses $\mK^l_s$ and $\mV^l_s$ in the GPU memory for inference on a dataset $\gD$, and obtains a similar result to a full model according to an evaluation scoring metric, i.e., $\text{score}(\mK^l, \mV^l, \gD) \approx \text{score}(\mK^l_s, \mV^l_s, \gD)$.

\subsection{Proposed Method}
\label{section:method}


In this section, we introduce our method, \method, based on the pyramidal information funneling observed across different layers in \Scref{section:observation}.  \method consists of two steps: (1) Dynamically allocating different KV cache sizes/budgets across different layers (\Scref{subsection:pyramid_size_allocation}); and (2) Selecting important KV vectors in each attention head for caching (\Scref{subsection:key_and_value_selection}).

\subsubsection{KV Cache Size/Budget Allocation}
\label{subsection:pyramid_size_allocation}
Previous work on KV cache compression~\citep{li2024snapkv,zhang2024h2o,xiao2023efficient} often allocates a fixed KV cache size across LLM layers. However, as our analysis in \Scref{section:observation} demonstrates, attention patterns are not identical across different layers. Particularly dense attention is observed in the lower layers, and sparse attention in higher layers. Therefore, using a fixed KV cache size across layers may lead to suboptimal performance. These approaches may retain many unimportant tokens in the higher layers of sparser attentions while potentially overlooking many crucial tokens in the lower layers of denser attentions.

Thus, we propose to increase compression efficiency by dynamically allocating the cache budgets across layers to reflect the aggregated information flow based on attention patterns.
Specifically, \method allocates more KV cache to the lower layers where information is
more dispersed and each KV state contains less information, while reducing the KV cache in higher layers where information becomes concentrated in a few key tokens.


Following the common practice in KV cache compression~\citep{li2024snapkv, xiao2023efficient}, 
we first retain the KV cache for the last $\alpha$ tokens of the input across all layers, as these tokens have been shown to contain the most immediate task-related information, where $\alpha$ is a hyperparameter, controlling the number of last few tokens being included in the KV cache.
For simplicity, we call these tokens ``\textit{instruction tokens}'', which is also referred to as ``\textit{local window}'' in previous literature~\citep{zhang2024h2o, li2024snapkv, xiao2023efficient}.  

Subsequently, given the remaining total cache budget $k^\text{total} = \sum_{l \in [0, m-1]} k^l$ that can be used over all  transformer layers (noted as $m$), we first determine the cache sizes for the top and bottom layers, and use an arithmetic sequence to compute the cache sizes for the intermediate layers to form the pyramidal shape. The key intuition is to follow the attention pattern in aggregated information flow, reflecting a monotonically decreasing pattern of important tokens for attention from lower layers to upper layers.
We allocate $k^{m-1} = k^\text{total} / (\beta \cdot m) $ for the top layer and $k^0 = (2 \cdot k^\text{total}) / m - k^{m-1}$ for the bottom layer,, where $\beta$ is a hyperparameter to adjust the pyramid's shape.
The hyperparameter $\beta$ is still required to determine the top layer. Once the top layer is identified, the budget of the bottom layer can be calculated by summing the budgets across all layers and equating this sum to the total budget.
Once the cache sizes of the bottom and top layers are determined, the cache sizes for all intermediate layers are set according to an arithmetic sequence, defined as 
\begin{align} \label{eq:budget}
k^l = k^{0} - \frac{k^{0} - k^{m-1}}{m - 1} \times l.
\end{align}




    

\subsubsection{KV Cache Selection}
\label{subsection:key_and_value_selection}

Once the KV cache budget is determined for each layer, our method needs to select specific KV states for caching within each layer in LLMs. As described in the previous section, the KV cache of the last $\alpha$ tokens, referred to as instruction tokens, are retained across all layers. Following SnapKV~\citep{li2024snapkv}, the selection of the remaining tokens is then guided by the attention scores derived from these instruction tokens---tokens receiving higher attention scores are deemed more relevant to the generation process and are thus their KV states are prioritized for retention in the GPU cache.

In a typical LLM, the attention mechanism in each head $h$ is calculated using the formula:  
\begin{align} \label{eq:Al}
    \mA^h = \text{softmax}(\mQ^h \cdot (\mK^h)^\top / \sqrt{d_k}),
\end{align}
where $d_k$ denotes the dimension of the key vectors. Following~\citep{li2024snapkv}, we utilize a pooling layer at $\mA^h$ to avoid the risk of being misled by some massive activation scores. 

To quantify the importance of each token during the generation process, we measure the level of attention each token receives from the instruction tokens, and use this measurement to select important tokens for KV caching. Specifically, we compute the score of selecting $i$-th token for retention in the KV cache as $s^h_i$ in each attention head $h$ by: 
\begin{align}\label{eq:att_score}
    s^h_i = \sum_{j\in [n-\alpha,n]} \mA^h_{ij}
\end{align}
where $[n-\alpha, n]$ is the range of the instruction tokens. In each layer $l$ and for each head $h$, the top $k^l$ tokens with the highest scores are selected, and their respective KV caches are retained. All other KV caches are discarded and will not be utilized in any subsequent computations throughout the generation process.

\section{Experiment}
\label{section:experiment}


We conduct comprehensive experiments to evaluate the effectiveness of \method on performance preserving and memory reduction.


\subsection{Experiment Setup}
\label{subsection:experiment_setup}

We maintain a fixed constant KV cache size for each layer for the baseline methods. In contrast, \method employs varying KV cache sizes across different layers. To ensure a fair comparison, we adjusted the average KV cache size in \method to match that of the baseline models, to keep the total memory consumption of all methods the same.
We set $\beta=20$  and $\alpha=8$. We use the same prompt for each dataset in all experiments.

\subsubsection{Backbone LLMs}
\label{subsubsection:llms}

We compare \method against baselines using state-of-the-art open-sourced LLMs, namely LLaMa-3-8B-Instruct, Mistral-7B-Instruct~\citep{jiang2023mistral} and  LLaMa-3-70B-Instruct. Testing examples are evaluated in a generative format, with answers generated by greedy decoding across all tasks to ensure a fair comparison.

\subsubsection{Datasets}
\label{subsubsection:datasets}

We use LongBench~\citep{bai2023longbench} to assess the performance of \method on tasks involving long-context inputs. LongBench is a meticulously designed benchmark suite that tests the capabilities of language models in handling extended documents and complex information sequences. This benchmark was created for comprehensive multi-task evaluation of long context inputs. It includes 17 datasets covering tasks such as single-document QA~\citep{kovcisky2018narrativeqa,dasigi2021dataset}, multi-document QA~\citep{yang2018hotpotqa,ho2020constructing}, summarization~\citep{huang2021efficient,zhong2021qmsum,fabbri2019multi}, few-shot learning~\citep{li2002learning,gliwa2019samsum,joshi2017triviaqa}, synthetic, and code generation~\citep{guo2023longcoder,liu2023repobench}. The datasets feature an average input length ranging from 1,235 to 18,409 tokens (detailed average lengths can be found in~\autoref{table:longbench}), necessitating substantial memory for KV cache management. For all these tasks, we adhered to the standard metrics recommended by LongBench~\citep{bai2023longbench} (i.e., F1 for QA, Rouge-L for summarization, Acc. for synthetic and Edit Sim. for code generation.) We refer readers to more details at~\autoref{appendix:evaluation}.

\subsubsection{Baselines}
\label{subsubsection:baselines}

We compare \method with three baselines, all of which keep the same KV cache size across different layers, with different strategies for KV cache selection.

\begin{itemize}
    \item \textbf{StreamingLLM (SLM)}~\citep{xiao2023efficient} is an efficient framework that enables LLMs to accept infinite input length.
    \item \textbf{Heavy Hitter Oracle (H2O)}~\citep{zhang2024h2o} is a KV cache compression policy that dynamically retains a balance of recent and Heavy Hitter (H2) tokens.
    \item \paragraph{SnapKV (SKV)}~\citep{li2024snapkv} automatically compresses KV caches by selecting clustered important tokens for each attention head.
    \item \paragraph{FullKV (FKV)} caches all keys and values for each input token in each layer. All methods are compared to the FullKV simultaneously.
\end{itemize}

\subsection{Main Results}
\label{subsection:results}


\begin{figure}
    \centering
    \begin{minipage}{0.33\textwidth}
        \centering
        \includegraphics[width=\linewidth]{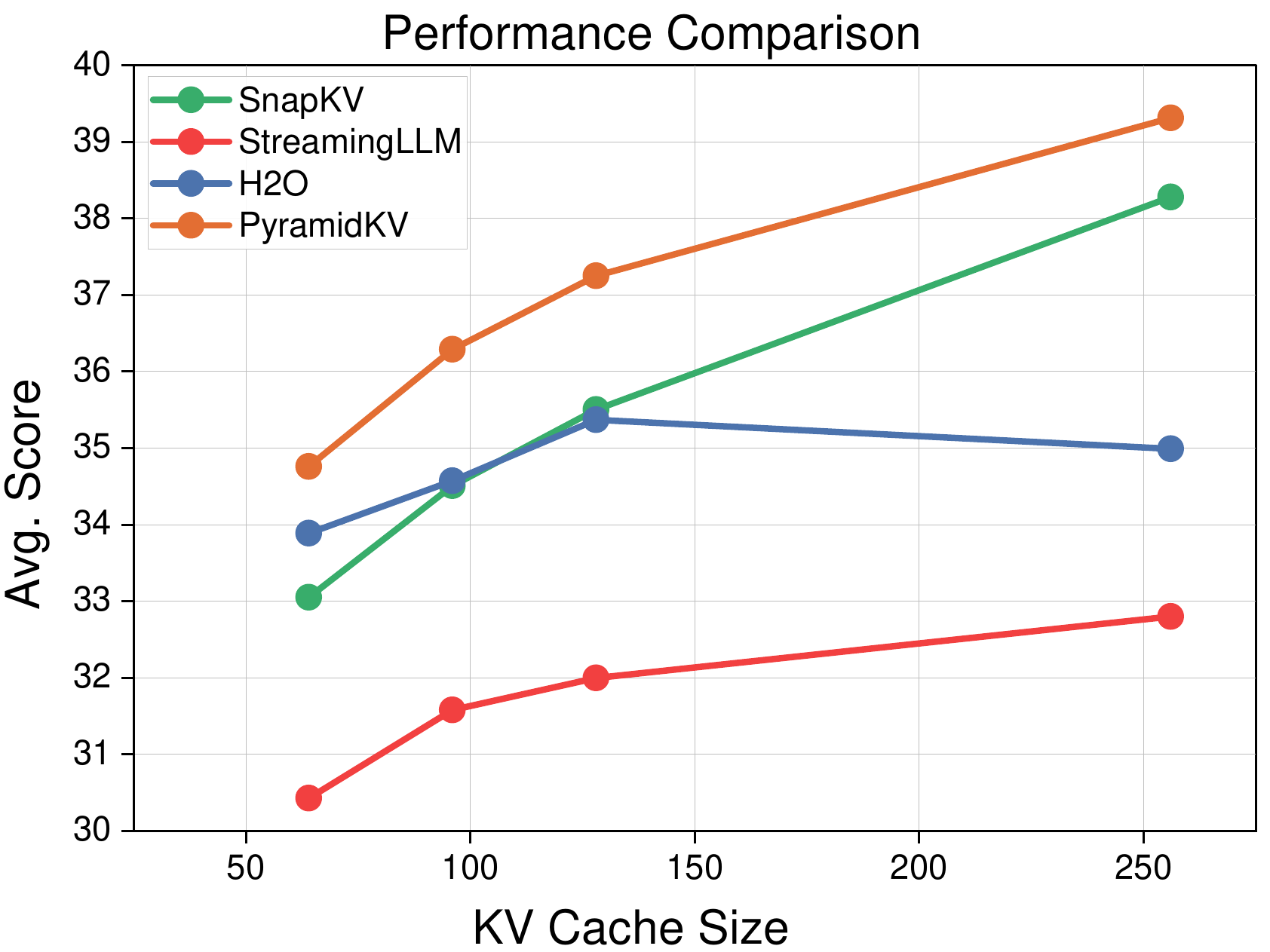}
    \end{minipage}%
    \hfill
    \begin{minipage}{0.33\textwidth}
        \centering
        \includegraphics[width=\linewidth]{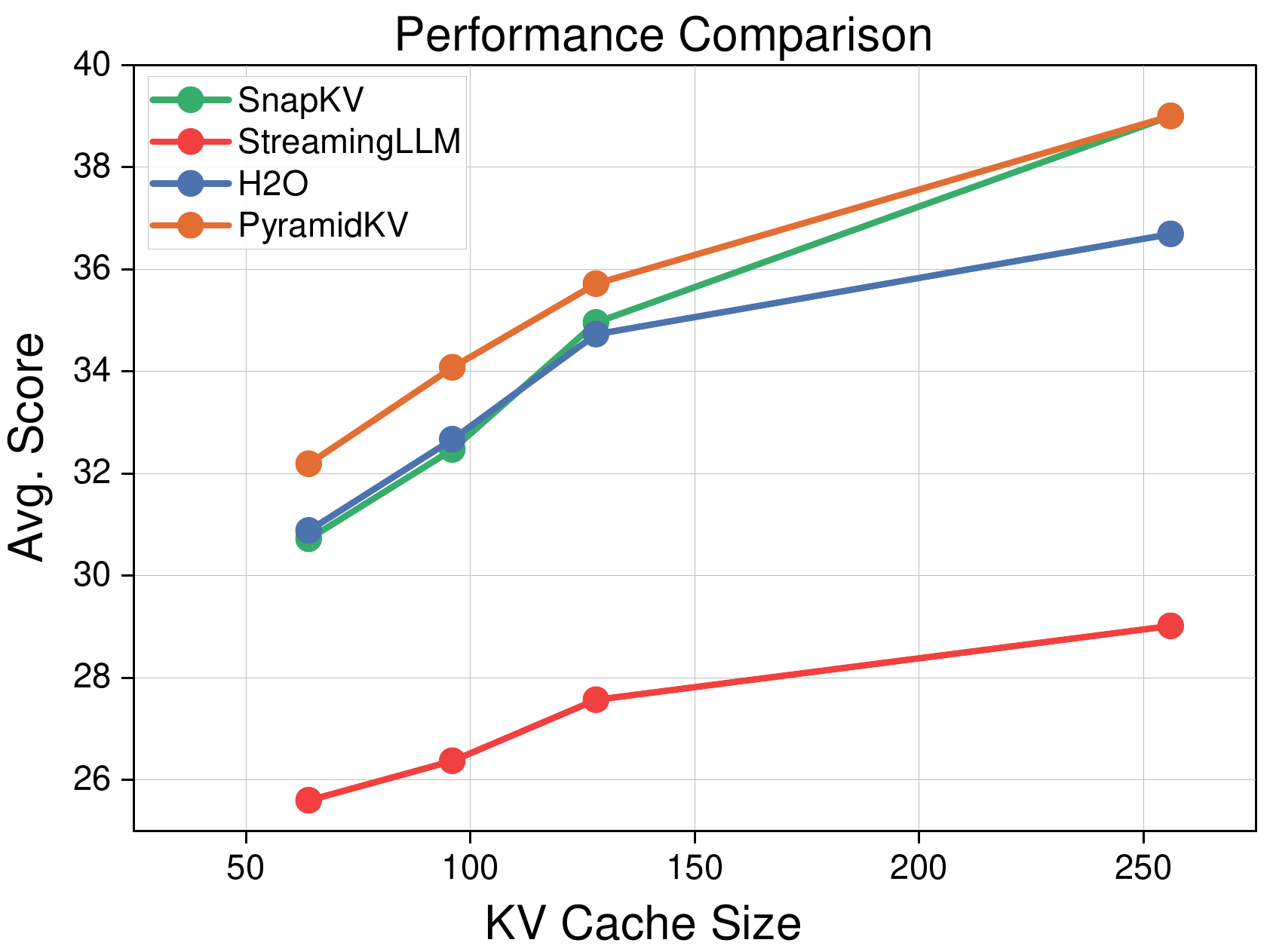}
    \end{minipage}%
    \hfill
    \begin{minipage}{0.33\textwidth}
        \centering
        \includegraphics[width=\linewidth]{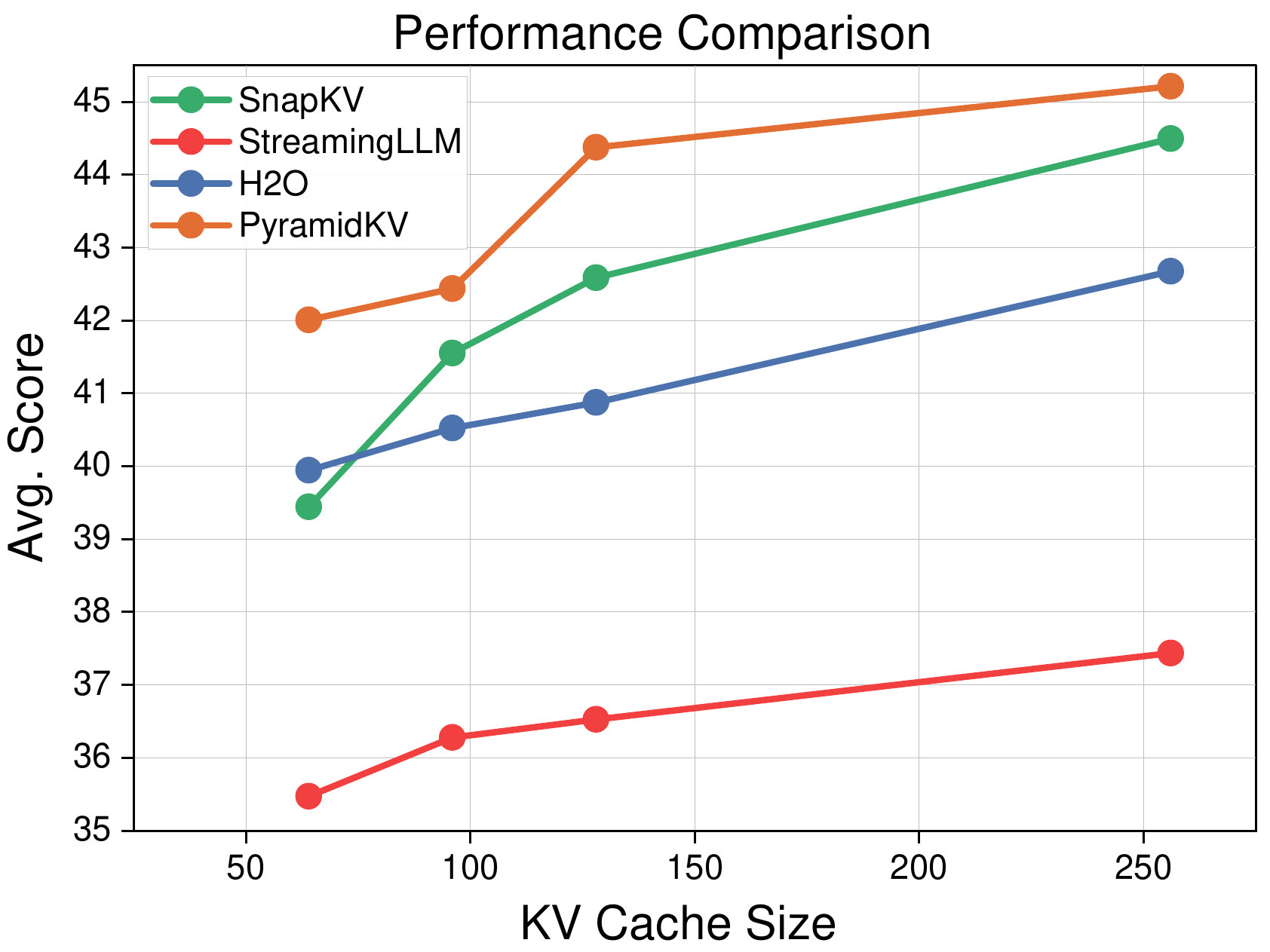}
    \end{minipage}
    \vspace{-3mm}
    \caption{The evaluation results from LongBench~\citep{bai2023longbench} across 64, 96, 128 and 256 cache sizes at LLaMa-3-8B-Instruct (Left), Mistral-7B-Instruct (Middle) and  LLaMa-3-70B-Instruct (Right).
    The evaluation metrics are the average score of LongBench across datasets.
    ~\method outperforms H2O~\citep{zhang2024h2o}, SnapKV~\citep{li2024snapkv} and StreamingLLM~\citep{xiao2023efficient}, especially in small KV cache sizes.}
    \label{figure:results}
\end{figure}

\begin{table*}[t]

\centering


\resizebox{\textwidth}{!}{
\begin{tabular}{l@{\hspace{0.05ex}}c@{\hspace{0.05ex}}c@{\hspace{0.05ex}}c@{\hspace{0.05ex}}c@{\hspace{0.05ex}}c@{\hspace{0.05ex}}c@{\hspace{0.05ex}}c@{\hspace{0.05ex}}c@{\hspace{0.05ex}}c@{\hspace{0.05ex}}c@{\hspace{0.05ex}}c@{\hspace{0.05ex}}c@{\hspace{0.05ex}}c@{\hspace{0.05ex}}c@{\hspace{0.6ex}}c@{\hspace{0.6ex}}c@{\hspace{0.6ex}}c}

\specialrule{1pt}{0pt}{2pt}
\multirow{5}{*}{Method}  & \multicolumn{3}{c}{Single-Document QA} & \multicolumn{3}{c}{Multi-Document QA}& \multicolumn{3}{c}{Summarization}& \multicolumn{3}{c}{Few-shot Learning}& \multicolumn{2}{c}{Synthetic} & \multicolumn{2}{c}{Code} & \multirow{5}{*}{Avg.} \\
\cmidrule(lr){2-4}\cmidrule(lr){5-7}\cmidrule(lr){8-10}\cmidrule(lr){11-13}\cmidrule(lr){14-15}\cmidrule(lr){16-17}
& \rotatebox[origin=c]{30}{NrtvQA} & \rotatebox[origin=c]{30}{Qasper} & \rotatebox[origin=c]{30}{MF-en} & \rotatebox[origin=c]{30}{HotpotQA} & \rotatebox[origin=c]{30}{2WikiMQA} & \rotatebox[origin=c]{30}{Musique} & \rotatebox[origin=c]{30}{GovReport} & \rotatebox[origin=c]{30}{QMSum} & \rotatebox[origin=c]{30}{MultiNews} & \rotatebox[origin=c]{30}{TREC} & \rotatebox[origin=c]{30}{TriviaQA} & \rotatebox[origin=c]{30}{SAMSum} & \rotatebox[origin=c]{30}{PCount} & \rotatebox[origin=c]{30}{PRe} & \rotatebox[origin=c]{30}{Lcc} & \rotatebox[origin=c]{30}{RB-P} & \\
\cmidrule(lr){2-17}
&18409&3619&4559&9151&4887&11214&8734&10614&2113&5177&8209&6258&11141&9289&1235&4206& \\

\midrule
\multicolumn{18}{c}{LlaMa-3-8B-Instruct, KV Size = Full} \\
\arrayrulecolor{black!20}\midrule
FKV &25.70 & 29.75 & 41.12 & 45.55 & 35.87 & 22.35 & 25.63 & 23.03 & 26.21 & 73.00 & 90.56 & 41.88 & 4.67 & 69.25 & 58.05 & 50.77 & 41.46 \\

\arrayrulecolor{black!20}\midrule
\multicolumn{18}{c}{LlaMa-3-8B-Instruct, KV Size = 64} \\
\arrayrulecolor{black!20}\midrule
SKV & 19.86 & 9.09 & 27.89 & \textbf{37.34} & 28.35 & \textbf{18.17} & 15.86 & 20.80 & 16.41 & 38.50 & 85.92 & 36.32 & 5.22 & 69.00 & 51.78 & 48.38 & 33.05 \\
H2O & 20.80 & 11.34 & 27.03 & 37.25 & 30.01 & 17.94 & 18.29 & 21.49 & 19.13 & 38.00 & 84.70 & \textbf{37.76} & \textbf{5.63} & 69.33 & \textbf{53.44} & \textbf{50.15} & 33.89 \\
SLM & 17.44 & 8.68 & 22.25 & 35.37 & \textbf{31.51} & 15.97 & 15.46 & 20.06 & 14.64 & 38.00 & 72.33 & 29.10 & 5.42 & \textbf{69.50} & 46.14 & 45.09 & 30.43 \\
Ours & \textbf{21.13} & \textbf{14.18} & \textbf{30.26} & 35.12 & 23.76 & 16.17 & \textbf{18.33} & \textbf{21.65} & \textbf{19.23} & \textbf{58.00} & \textbf{88.31} & 37.07 & 5.23 & \textbf{69.50} & 52.61 & 45.74 & \textbf{34.76} \\


\arrayrulecolor{black!20}\midrule
\multicolumn{18}{c}{LlaMa-3-8B-Instruct, KV Size = 2048} \\
\arrayrulecolor{black!20}\midrule
SKV & \textbf{25.86} & 29.55 & \textbf{41.10} & \textbf{44.99} & \textbf{35.80} & 21.81 & 25.98 & \textbf{23.40} & 26.46 & \textbf{73.50} & \textbf{90.56} & 41.66 & 5.17 & \textbf{69.25} & 56.65 & 49.94 & 41.35 \\
SLM & 21.71 & 25.78 & 38.13 & 40.12 & 32.01 & 16.86 & 23.14 & 22.64 & \textbf{26.48} & 70.00 & 83.22 & 31.75 & \textbf{5.74} & 68.50 & 53.50 & 45.58 & 37.82 \\
H2O & 25.56 & 26.85 & 39.54 & 44.30 & 32.92 & 21.09 & 24.68 & 23.01 & 26.16 & 53.00 & \textbf{90.56} & 41.84 & 4.91 & \textbf{69.25} & 56.40 & 49.68 & 39.35 \\
Ours & 25.40 & \textbf{29.71} & 40.25 & 44.76 & 35.32 & \textbf{21.98} & \textbf{26.83} & 23.30 & 26.19 & 73.00 & \textbf{90.56} & \textbf{42.14} & 5.22 & \textbf{69.25} & \textbf{58.76} & \textbf{51.18} & \textbf{41.49} \\

\arrayrulecolor{black}\midrule
\multicolumn{18}{c}{Mistral-7B-Instruct, KV Size = Full} \\
\arrayrulecolor{black!20}\midrule

FKV & 26.90 & 33.07 & 49.20 & 43.02 & 27.33 & 18.78 & 32.91 & 24.21 & 26.99 & 71.00 & 86.23 & 42.65 & 2.75 & 86.98 & 56.96 & 54.52 & 42.71 \\

\arrayrulecolor{black!20}\midrule
\multicolumn{18}{c}{Mistral-7B-Instruct, KV Size = 64} \\
\arrayrulecolor{black!20}\midrule
SKV & 16.94&17.17&39.51&\textbf{36.87}&22.26&15.18&14.75&20.35&21.45&37.50&\textbf{84.16}&37.28&4.50&61.13&42.40&38.44&30.72\\
SLM & 15.01&13.84&28.74&30.97&\textbf{24.50}&13.42&13.25&19.46&19.17&35.50&76.91&29.61&4.67&27.33&38.71&35.29&25.60\\
H2O & 18.19&19.04&37.40&30.18&22.22&13.77&16.60&\textbf{21.52}&\textbf{21.98}&37.00&81.02&\textbf{38.62}&\textbf{5.00}&\textbf{66.03}&43.54&\textbf{40.46}&30.88\\
Ours & \textbf{20.91}&\textbf{20.21}&\textbf{39.94}&33.57&22.87&\textbf{15.70}&\textbf{17.31}&21.23&21.41&\textbf{54.00}&81.98&36.96&3.58&60.83&\textbf{44.52}&37.99&\textbf{32.19}\\


\arrayrulecolor{black!20}\midrule
\multicolumn{18}{c}{Mistral-7B-Instruct, KV Size = 2048} \\
\arrayrulecolor{black!20}\midrule

SKV &\textbf{25.89} & \textbf{32.93} & 48.56 & \textbf{42.96} & 27.42 & 19.02 & 26.56 & \textbf{24.47} & 26.69 & 70.00 & 86.27 & 42.57 & \textbf{5.50} & \textbf{88.90} & 50.42 & 46.72 & 41.56 \\
SLM & 20.31 & 26.64 & 45.72 & 35.25 & 24.31 & 12.20 & \textbf{27.47} & 21.57 & 24.51 & 68.50 & 71.95 & 31.19 & 5.00 & 22.56 & 43.38 & 37.08 & 32.35 \\
H2O & 25.76 & 31.10 & \textbf{49.03} & 40.76 & 26.52 & 17.07 & 24.81 & 23.64 & 26.60 & 55.00 & \textbf{86.35} & 42.48 & \textbf{5.50} & 88.15 & 49.93 & 46.57 & 39.95 \\
Ours & 25.53 & 32.21 & 48.97 & 42.26 & \textbf{27.50} & \textbf{19.36} & 26.60 & 23.97 & \textbf{26.73} & \textbf{71.00} & 86.25 & \textbf{42.94} & 4.50 & 87.90 & \textbf{53.12} & \textbf{47.21} & \textbf{41.63} \\

\arrayrulecolor{black}\midrule
\multicolumn{18}{c}{LlaMa-3-70B-Instruct, KV Size = Full} \\
\arrayrulecolor{black!20}\midrule
FKV & 27.75 & 46.48 & 49.45 & 52.04 & 54.90 & 30.42 & 32.37 & 22.27 & 27.58 & 73.50 & 92.46 & 45.73 & 12.50 & 72.50 & 40.96 & 63.91 & 46.55\\

\arrayrulecolor{black!20}\midrule
\multicolumn{18}{c}{LlaMa-3-70B-Instruct, KV Size = 64} \\
\arrayrulecolor{black!20}\midrule
SKV & 23.92&31.09&36.54&46.66&50.40&25.30&18.05&21.11&19.79&41.50&\textbf{91.06}&40.26&12.00&\textbf{72.50}&43.33&57.62&39.45\\
SLM & 22.07&23.53&27.31&43.21&\textbf{51.66}&23.85&16.62&19.74&15.20&39.50&76.89&33.06&12.00&\textbf{72.50}&40.23&50.20&35.47\\
H2O & 25.45&34.64&33.23&\textbf{48.25}&50.30&24.88&20.03&21.50&21.39&42.00&90.36&\textbf{41.58}&12.00&71.50&43.83&\textbf{58.16}&39.94\\
Ours & \textbf{25.47}&\textbf{36.71}&\textbf{42.29}&47.08&46.21&\textbf{28.30}&\textbf{20.60}&\textbf{21.62}&\textbf{21.62}&\textbf{64.50}&89.61&41.28&\textbf{12.50}&\textbf{72.50}&\textbf{45.34}&56.50&\textbf{42.01}\\


\arrayrulecolor{black!20}\midrule
\multicolumn{18}{c}{LlaMa-3-70B-Instruct, KV Size = 2048} \\
\arrayrulecolor{black!20}\midrule
SKV & 26.73&45.18&47.91&\textbf{52.00}&\textbf{55.24}&30.48&28.76&22.35&27.31&72.50&92.38&45.58&12.00&\textbf{72.50}&\textbf{41.52}&\textbf{69.27}&46.36\\ 
SLM & 26.69&41.01&35.97&46.55&52.98&25.71&27.81&20.81&27.16&69.00&91.55&44.02&12.00&72.00&41.44&68.73&43.96\\ 
H2O & \textbf{27.67}&\textbf{46.51}&\textbf{49.54}&51.49&53.85&29.97&28.57&\textbf{22.79}&\textbf{27.53}&59.00&\textbf{92.63}&\textbf{45.94}&12.00&72.50&41.39&63.90&45.33\\ 
Ours & 27.22&46.19&48.72&51.62&54.56&\textbf{31.11}&\textbf{29.76}&22.50&27.27&\textbf{73.50}&91.88&45.47&12.00&72.50&41.36&69.12&\textbf{46.55}\\

\arrayrulecolor{black}\bottomrule
\end{tabular}
}
\caption{Performance comparison of ~\method (Ours) with SnapKV (SKV), H2O, StreamingLLM (SLM) and FullKV (FKV) on LongBench for LlaMa-3-8B-Instruct, Mistral-7B-Instruct and LlaMa-3-70B-Instruct. ~\method generally outperforms other KV Cache compression methods across various KV Cache sizes and LLMs. The performance strengths of ~\method are more evident in small KV Cache sizes (i.e. KV Size = 64).
}

\label{table:longbench}
\vspace{-3mm}
\end{table*}

The evaluation results from LongBench~\citep{bai2023longbench} are shown in \autoref{table:longbench} and \autoref{figure:results}.
In \autoref{figure:results}, we report the average score across datasets for 64, 96, 128, and 256 case sizes.
In \autoref{table:longbench}, we report the results for two different KV cache sizes with 64 and 2048. These two sizes represent two distinct operational scenarios---the memory-efficient scenario and the performance-preserving scenario, respectively for a trade-off between memory and model performance. 
In \autoref{appendix:results}, we report results of KV cache sizes with 64, 96, 128 and 2048.

\begin{figure}[t]
    \centering
    \includegraphics[width=\columnwidth]{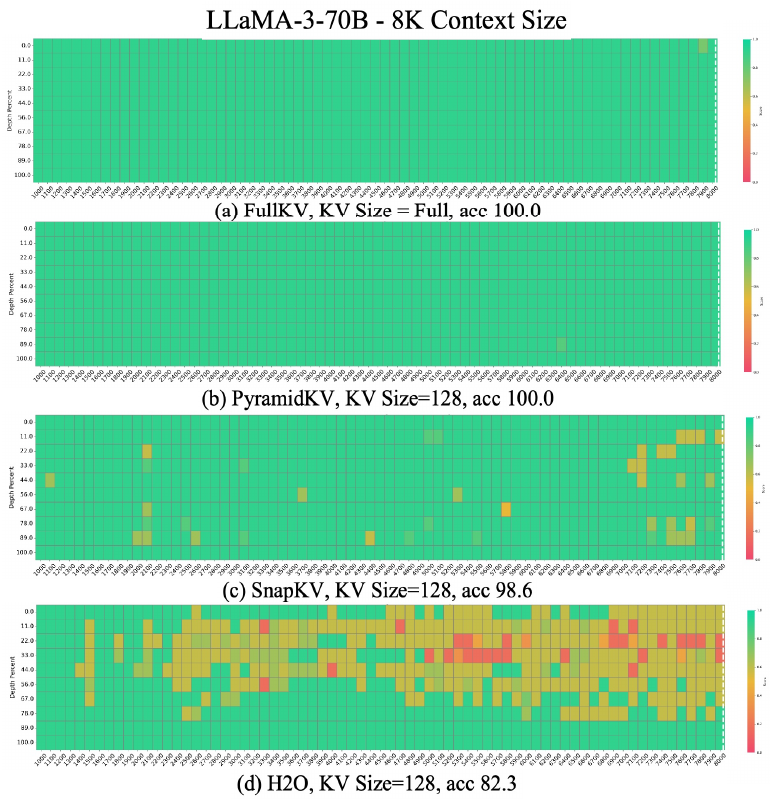}
    \vspace{-0.5cm}
    \caption{
    Results of the Fact Retrieval Across Context Lengths (``Needle In A HayStack'') test in \textbf{LlaMa-3-70B-Instruct} with \textbf{8k} context size in \textbf{128} KV cache size.
    The vertical axis of the table represents the depth percentage, and the horizontal axis represents the length.
    }
    \label{figure:needle_llama}
\end{figure}

Overall, \method preserves the performance with only 12\% of the KV cache and it consistently surpasses other method across a range of KV cache sizes and different backbone models, with its performance advantages becoming particularly pronounced in memory-constrained environments where only about 0.8\% of the KV cache from the prompt is retained. Upon examining specific tasks, \method demonstrates a notably superior performance on the TREC task, a few-shot question answering challenge. This suggests that the model effectively aggregates information from the few-shot examples, highlighting the potential for further investigation into in-context learning tasks.

Notably, we initially observe the pyramidal attention patterns from the visualization analysis on the multi-document QA task (\autoref{figure:rag_flow}), but the pyramid heuristic has demonstrated its effectiveness on a range of other LongBench tasks (e.g., single-document QA, In-Context Learning), suggesting its promising generalizability beyond multi-document QA.

The performance advantage of \method increases as the KV cache memory decreases. By focusing on optimizing budget allocation across layers, \method accurately allocates resources in memory-constrained scenarios, ensuring that retained information is effectively preserved to maintain model performance. Moreover, as in long bench results shown in \autoref{table:longbench}, even in the performance-preserving scenario (i.e., KV cache size = 2048), \method improves the performance over baseline methods and even outperforms FullKV.

Among the 16 datasets, the tasks where our proposed method performs slightly worse than the baseline are mostly saturated (e.g., HotpotQA, Musique, etc under the LlaMa-3-8B-Instruct setting with KV Size = 64, as shown in Table 1). In these cases, our method is only marginally inferior to the baseline and remains competitive.
Conversely, on tasks with greater potential for improvement (e.g., Qasper, MF-en, TREC, TriviaQA, etc under the same setting), our method significantly outperforms the baseline. Consequently, the overall average performance of our method surpasses that of the baselines. Notably, these tasks include several In-Context Learning tasks (i.e., TREC), our method enjoys best performance gain at In-Context Learning tasks.


\subsection{Discussion and Insights}
\label{subsection:insight_experiment}
\subsubsection{\method Preserves the Long-Context Understanding Ability}
\label{subsubsection:needle_in_the_haystack}



We conduct the "Fact Retrieval Across Context Lengths" (Needle In A Haystack) experiment~\citep{liu2023lost,fu2024data}, which is a dataset designed to test whether a model can find key information in long input sequences, to evaluate the in-context retrieval capabilities of LLMs when utilizing various KV cache compression methods. For this purpose, we employ \textbf{LlaMa-3-70B-Instruct} as our base, with context lengths extending up to 8k. We compared several KV cache compression techniques (\method, SnapKV~\citep{li2024snapkv}, and H2O~\citep{zhang2024h2o}) at cache sizes of \textbf{128} and full cache. The results, presented in \autoref{figure:needle_llama} \footnote{Additional results with 64, 96 and 128 KV cache sizes with \textbf{LlaMa-3-8B-Instruct} at 8k context length, \textbf{LlaMa-3-70B-Instruct} at 8k context length, and \textbf{Mistral-7B-Instruct}~\citep{jiang2023mistral} at 32k context length are available in \autoref{appendix:needle}}. The results demonstrate that with only 128 KV cache retained, \method effectively maintains the model's ability to understand short contexts, and shows only modest degradation for longer contexts. In contrast, other KV cache compression methods significantly hinder the performance of LLMs. Notably, for the larger model (\textbf{LlaMa-3-70B-Instruct}), \method achieves 100.0 Acc. performance, matching the results of FullKV, thereby demonstrating its ability to preserve long-context comprehension with a substantially reduced KV cache.
We adopt the haystack setting of haystack formed from a long corpus for the Needle In A Haystack task as \cite{wu2024retrieval}.

\subsubsection{~\method  Significantly Reduces Memory with Limited Performance Drop}
\label{subsubsection:memory}


In this section, we study how sensitive the methods are with different sizes of KV cache. 
We report the KV cache memory reduction in Table \ref{tab:memory_reduction}.
We evaluate the memory consumption of LLaMa-3-8B-Instruct.
Specifically, we evaluate the memory consumption of all methods with a fixed batch size of 1, a sequence length of 8192, and model weights in fp16 format.
We observe that \method substantially reduces the KV cache memory across different numbers of cache sizes.
We also present that the allocation strategy and score-based selection add minimal complexity in the inference phase as~\autoref{appendix:inference_overhead}.

\begin{table*}[h]
\small
\centering

\resizebox{\textwidth}{!}{
\begin{tabular}{ccc|cccccc}
\toprule
 cache size & Memory & Compression Ratio & QMSum & TREC & TriviaQA & PCount & PRe & Lcc \\

\midrule
512 & 428M & 6.3\% & 22.80 & 71.50 & 90.61 & 5.91 & 69.50 & 58.16 \\
1024 & 856M & 12.5\% & 22.55 & 71.50 & 90.61 & 5.91 & 69.50 & 58.16 \\
2048 & 1712M & 25.0\% & 22.55 & 72.00 & 90.56 & 5.58 & 69.25 & 56.79 \\
Full & 6848M & 100.0\% & 23.30 & 73.00 & 90.56 & 5.22 & 69.25 & 58.76 \\

\bottomrule
\end{tabular}
}
\caption{Memory reduction effect and benchmark result by using \method. We conducted a comparison of memory consumption between the Llama-3-8B-Instruct model utilizing the Full KV cache and the Llama-3-8B-Instruct model compressed with the \method.
}

\label{tab:memory_reduction}
\vspace{-3mm}
\end{table*}

\section{Conclusion}
\label{section:conclusion}
In this study, we investigate Pyramidal Information Funneling, the intrinsic attention patterns of Large Language Models (LLMs) when processing long context inputs.
Motivated by this discovery, we design a novel KV cache compression approach \method that utilizes this information flow pattern.
Our method excels in memory-constrained settings, preserves long-context understanding ability, and significantly reduces memory usage.

\bibliographystyle{plainnat}
\bibliography{reference}

\begin{thebibliography}{40}
\providecommand{\natexlab}[1]{#1}
\providecommand{\url}[1]{\texttt{#1}}
\expandafter\ifx\csname urlstyle\endcsname\relax
  \providecommand{\doi}[1]{doi: #1}\else
  \providecommand{\doi}{doi: \begingroup \urlstyle{rm}\Url}\fi

\bibitem[Achiam et~al.(2023)Achiam, Adler, Agarwal, Ahmad, Akkaya, Aleman, Almeida, Altenschmidt, Altman, Anadkat, et~al.]{achiam2023gpt}
Josh Achiam, Steven Adler, Sandhini Agarwal, Lama Ahmad, Ilge Akkaya, Florencia~Leoni Aleman, Diogo Almeida, Janko Altenschmidt, Sam Altman, Shyamal Anadkat, et~al.
\newblock Gpt-4 technical report.
\newblock \emph{arXiv preprint arXiv:2303.08774}, 2023.

\bibitem[Bai et~al.(2023)Bai, Lv, Zhang, Lyu, Tang, Huang, Du, Liu, Zeng, Hou, et~al.]{bai2023longbench}
Yushi Bai, Xin Lv, Jiajie Zhang, Hongchang Lyu, Jiankai Tang, Zhidian Huang, Zhengxiao Du, Xiao Liu, Aohan Zeng, Lei Hou, et~al.
\newblock Longbench: A bilingual, multitask benchmark for long context understanding.
\newblock \emph{arXiv preprint arXiv:2308.14508}, 2023.

\bibitem[Chen et~al.(2024{\natexlab{a}})Chen, Zhao, Liu, Bai, Lin, Zhou, and Chang]{chen2024image}
Liang Chen, Haozhe Zhao, Tianyu Liu, Shuai Bai, Junyang Lin, Chang Zhou, and Baobao Chang.
\newblock An image is worth 1/2 tokens after layer 2: Plug-and-play inference acceleration for large vision-language models.
\newblock \emph{arXiv preprint arXiv:2403.06764}, 2024{\natexlab{a}}.

\bibitem[Chen et~al.(2023)Chen, Qian, Tang, Lai, Liu, Han, and Jia]{chen2023longlora}
Yukang Chen, Shengju Qian, Haotian Tang, Xin Lai, Zhijian Liu, Song Han, and Jiaya Jia.
\newblock Longlora: Efficient fine-tuning of long-context large language models.
\newblock \emph{arXiv preprint arXiv:2309.12307}, 2023.

\bibitem[Chen et~al.(2024{\natexlab{b}})Chen, Sadhukhan, Ye, Zhou, Zhang, Nolte, Tian, Douze, Bottou, Jia, and Chen]{chen2024magicpiglshsamplingefficient}
Zhuoming Chen, Ranajoy Sadhukhan, Zihao Ye, Yang Zhou, Jianyu Zhang, Niklas Nolte, Yuandong Tian, Matthijs Douze, Leon Bottou, Zhihao Jia, and Beidi Chen.
\newblock Magicpig: Lsh sampling for efficient llm generation, 2024{\natexlab{b}}.
\newblock URL \url{https://arxiv.org/abs/2410.16179}.

\bibitem[Chiang et~al.(2023)Chiang, Li, Lin, Sheng, Wu, Zhang, Zheng, Zhuang, Zhuang, Gonzalez, Stoica, and Xing]{vicuna2023}
Wei-Lin Chiang, Zhuohan Li, Zi~Lin, Ying Sheng, Zhanghao Wu, Hao Zhang, Lianmin Zheng, Siyuan Zhuang, Yonghao Zhuang, Joseph~E. Gonzalez, Ion Stoica, and Eric~P. Xing.
\newblock Vicuna: An open-source chatbot impressing gpt-4 with 90\%* chatgpt quality, March 2023.
\newblock URL \url{https://lmsys.org/blog/2023-03-30-vicuna/}.

\bibitem[Dasigi et~al.(2021)Dasigi, Lo, Beltagy, Cohan, Smith, and Gardner]{dasigi2021dataset}
Pradeep Dasigi, Kyle Lo, Iz~Beltagy, Arman Cohan, Noah~A Smith, and Matt Gardner.
\newblock A dataset of information-seeking questions and answers anchored in research papers.
\newblock In \emph{Proceedings of the 2021 Conference of the North American Chapter of the Association for Computational Linguistics: Human Language Technologies}, pages 4599--4610, 2021.

\bibitem[Ding et~al.(2024)Ding, Zhang, Zhang, Xu, Shang, Xu, Yang, and Yang]{ding2024longrope}
Yiran Ding, Li~Lyna Zhang, Chengruidong Zhang, Yuanyuan Xu, Ning Shang, Jiahang Xu, Fan Yang, and Mao Yang.
\newblock Longrope: Extending llm context window beyond 2 million tokens.
\newblock \emph{arXiv preprint arXiv:2402.13753}, 2024.

\bibitem[Dong et~al.(2024)Dong, Yang, Zhang, Wang, Chi, and Chen]{dong2024get}
Harry Dong, Xinyu Yang, Zhenyu Zhang, Zhangyang Wang, Yuejie Chi, and Beidi Chen.
\newblock Get more with less: Synthesizing recurrence with kv cache compression for efficient llm inference.
\newblock \emph{arXiv preprint arXiv:2402.09398}, 2024.

\bibitem[Fabbri et~al.(2019{\natexlab{a}})Fabbri, Li, She, Li, and Radev]{fabbri-etal-2019-multi}
Alexander Fabbri, Irene Li, Tianwei She, Suyi Li, and Dragomir Radev.
\newblock Multi-news: A large-scale multi-document summarization dataset and abstractive hierarchical model.
\newblock In Anna Korhonen, David Traum, and Llu{\'\i}s M{\`a}rquez, editors, \emph{Proceedings of the 57th Annual Meeting of the Association for Computational Linguistics}, pages 1074--1084, Florence, Italy, July 2019{\natexlab{a}}. Association for Computational Linguistics.
\newblock \doi{10.18653/v1/P19-1102}.
\newblock URL \url{https://aclanthology.org/P19-1102}.

\bibitem[Fabbri et~al.(2019{\natexlab{b}})Fabbri, Li, She, Li, and Radev]{fabbri2019multi}
Alexander~Richard Fabbri, Irene Li, Tianwei She, Suyi Li, and Dragomir Radev.
\newblock Multi-news: A large-scale multi-document summarization dataset and abstractive hierarchical model.
\newblock In \emph{Proceedings of the 57th Annual Meeting of the Association for Computational Linguistics}, pages 1074--1084, 2019{\natexlab{b}}.

\bibitem[Fu et~al.(2024)Fu, Panda, Niu, Yue, Hajishirzi, Kim, and Peng]{fu2024data}
Yao Fu, Rameswar Panda, Xinyao Niu, Xiang Yue, Hannaneh Hajishirzi, Yoon Kim, and Hao Peng.
\newblock Data engineering for scaling language models to 128k context.
\newblock \emph{arXiv preprint arXiv:2402.10171}, 2024.

\bibitem[Ge et~al.(2023)Ge, Zhang, Liu, Zhang, Han, and Gao]{ge2023model}
Suyu Ge, Yunan Zhang, Liyuan Liu, Minjia Zhang, Jiawei Han, and Jianfeng Gao.
\newblock Model tells you what to discard: Adaptive kv cache compression for llms.
\newblock \emph{arXiv preprint arXiv:2310.01801}, 2023.

\bibitem[Gliwa et~al.(2019)Gliwa, Mochol, Biesek, and Wawer]{gliwa2019samsum}
Bogdan Gliwa, Iwona Mochol, Maciej Biesek, and Aleksander Wawer.
\newblock Samsum corpus: A human-annotated dialogue dataset for abstractive summarization.
\newblock \emph{EMNLP-IJCNLP 2019}, page~70, 2019.

\bibitem[Guo et~al.(2023)Guo, Xu, Duan, Yin, and McAuley]{guo2023longcoder}
Daya Guo, Canwen Xu, Nan Duan, Jian Yin, and Julian McAuley.
\newblock Longcoder: A long-range pre-trained language model for code completion.
\newblock \emph{arXiv preprint arXiv:2306.14893}, 2023.

\bibitem[Han et~al.(2023)Han, Wang, Xiong, Chen, Ji, and Wang]{han2023lm}
Chi Han, Qifan Wang, Wenhan Xiong, Yu~Chen, Heng Ji, and Sinong Wang.
\newblock Lm-infinite: Simple on-the-fly length generalization for large language models.
\newblock \emph{arXiv preprint arXiv:2308.16137}, 2023.

\bibitem[Ho et~al.(2020)Ho, Nguyen, Sugawara, and Aizawa]{ho2020constructing}
Xanh Ho, Anh-Khoa~Duong Nguyen, Saku Sugawara, and Akiko Aizawa.
\newblock Constructing a multi-hop qa dataset for comprehensive evaluation of reasoning steps.
\newblock In \emph{Proceedings of the 28th International Conference on Computational Linguistics}, pages 6609--6625, 2020.

\bibitem[Huang et~al.(2021)Huang, Cao, Parulian, Ji, and Wang]{huang2021efficient}
Luyang Huang, Shuyang Cao, Nikolaus Parulian, Heng Ji, and Lu~Wang.
\newblock Efficient attentions for long document summarization.
\newblock In \emph{Proceedings of the 2021 Conference of the North American Chapter of the Association for Computational Linguistics: Human Language Technologies}, pages 1419--1436, 2021.

\bibitem[Jiang et~al.(2023)Jiang, Sablayrolles, Mensch, Bamford, Chaplot, Casas, Bressand, Lengyel, Lample, Saulnier, et~al.]{jiang2023mistral}
Albert~Q Jiang, Alexandre Sablayrolles, Arthur Mensch, Chris Bamford, Devendra~Singh Chaplot, Diego de~las Casas, Florian Bressand, Gianna Lengyel, Guillaume Lample, Lucile Saulnier, et~al.
\newblock Mistral 7b.
\newblock \emph{arXiv preprint arXiv:2310.06825}, 2023.

\bibitem[Jiang et~al.(2024)Jiang, Li, Zhang, Wu, Luo, Ahn, Han, Abdi, Li, Lin, et~al.]{jiang2024minference}
Huiqiang Jiang, Yucheng Li, Chengruidong Zhang, Qianhui Wu, Xufang Luo, Surin Ahn, Zhenhua Han, Amir~H Abdi, Dongsheng Li, Chin-Yew Lin, et~al.
\newblock Minference 1.0: Accelerating pre-filling for long-context llms via dynamic sparse attention.
\newblock \emph{arXiv preprint arXiv:2407.02490}, 2024.

\bibitem[Joshi et~al.(2017)Joshi, Choi, Weld, and Zettlemoyer]{joshi2017triviaqa}
Mandar Joshi, Eunsol Choi, Daniel~S Weld, and Luke Zettlemoyer.
\newblock Triviaqa: A large scale distantly supervised challenge dataset for reading comprehension.
\newblock In \emph{Proceedings of the 55th Annual Meeting of the Association for Computational Linguistics (Volume 1: Long Papers)}, pages 1601--1611, 2017.

\bibitem[Ko{\v{c}}isk{\`y} et~al.(2018)Ko{\v{c}}isk{\`y}, Schwarz, Blunsom, Dyer, Hermann, Melis, and Grefenstette]{kovcisky2018narrativeqa}
Tom{\'a}{\v{s}} Ko{\v{c}}isk{\`y}, Jonathan Schwarz, Phil Blunsom, Chris Dyer, Karl~Moritz Hermann, G{\'a}bor Melis, and Edward Grefenstette.
\newblock The narrativeqa reading comprehension challenge.
\newblock \emph{Transactions of the Association for Computational Linguistics}, 6:\penalty0 317--328, 2018.

\bibitem[Lee et~al.(2024)Lee, Lee, Seo, and Sim]{lee2024infinigenefficientgenerativeinference}
Wonbeom Lee, Jungi Lee, Junghwan Seo, and Jaewoong Sim.
\newblock Infinigen: Efficient generative inference of large language models with dynamic kv cache management, 2024.
\newblock URL \url{https://arxiv.org/abs/2406.19707}.

\bibitem[Li and Roth(2002)]{li2002learning}
Xin Li and Dan Roth.
\newblock Learning question classifiers.
\newblock In \emph{COLING 2002: The 19th International Conference on Computational Linguistics}, 2002.

\bibitem[Li et~al.(2024)Li, Huang, Yang, Venkitesh, Locatelli, Ye, Cai, Lewis, and Chen]{li2024snapkv}
Yuhong Li, Yingbing Huang, Bowen Yang, Bharat Venkitesh, Acyr Locatelli, Hanchen Ye, Tianle Cai, Patrick Lewis, and Deming Chen.
\newblock Snapkv: Llm knows what you are looking for before generation.
\newblock \emph{arXiv preprint arXiv:2404.14469}, 2024.

\bibitem[Liu et~al.(2023{\natexlab{a}})Liu, Lin, Hewitt, Paranjape, Bevilacqua, Petroni, and Liang]{liu2023lost}
Nelson~F. Liu, Kevin Lin, John Hewitt, Ashwin Paranjape, Michele Bevilacqua, Fabio Petroni, and Percy Liang.
\newblock Lost in the middle: How language models use long contexts, 2023{\natexlab{a}}.

\bibitem[Liu et~al.(2023{\natexlab{b}})Liu, Xu, and McAuley]{liu2023repobench}
Tianyang Liu, Canwen Xu, and Julian McAuley.
\newblock Repobench: Benchmarking repository-level code auto-completion systems, 2023{\natexlab{b}}.

\bibitem[Roziere et~al.(2023)Roziere, Gehring, Gloeckle, Sootla, Gat, Tan, Adi, Liu, Remez, Rapin, et~al.]{roziere2023code}
Baptiste Roziere, Jonas Gehring, Fabian Gloeckle, Sten Sootla, Itai Gat, Xiaoqing~Ellen Tan, Yossi Adi, Jingyu Liu, Tal Remez, J{\'e}r{\'e}my Rapin, et~al.
\newblock Code llama: Open foundation models for code.
\newblock \emph{arXiv preprint arXiv:2308.12950}, 2023.

\bibitem[Sun et~al.(2024)Sun, Chen, Kolter, and Liu]{sun2024massive}
Mingjie Sun, Xinlei Chen, J~Zico Kolter, and Zhuang Liu.
\newblock Massive activations in large language models.
\newblock \emph{arXiv preprint arXiv:2402.17762}, 2024.

\bibitem[Touvron et~al.(2023{\natexlab{a}})Touvron, Lavril, Izacard, Martinet, Lachaux, Lacroix, Rozi{\`e}re, Goyal, Hambro, Azhar, et~al.]{touvron2023llama}
Hugo Touvron, Thibaut Lavril, Gautier Izacard, Xavier Martinet, Marie-Anne Lachaux, Timoth{\'e}e Lacroix, Baptiste Rozi{\`e}re, Naman Goyal, Eric Hambro, Faisal Azhar, et~al.
\newblock Llama: Open and efficient foundation language models.
\newblock \emph{arXiv preprint arXiv:2302.13971}, 2023{\natexlab{a}}.

\bibitem[Touvron et~al.(2023{\natexlab{b}})Touvron, Martin, Stone, Albert, Almahairi, Babaei, Bashlykov, Batra, Bhargava, Bhosale, et~al.]{touvron2023llama2}
Hugo Touvron, Louis Martin, Kevin Stone, Peter Albert, Amjad Almahairi, Yasmine Babaei, Nikolay Bashlykov, Soumya Batra, Prajjwal Bhargava, Shruti Bhosale, et~al.
\newblock Llama 2: Open foundation and fine-tuned chat models.
\newblock \emph{arXiv preprint arXiv:2307.09288}, 2023{\natexlab{b}}.

\bibitem[Waddington et~al.(2013)Waddington, Colmenares, Kuang, and Song]{waddington2013kv}
Daniel Waddington, Juan Colmenares, Jilong Kuang, and Fengguang Song.
\newblock Kv-cache: A scalable high-performance web-object cache for manycore.
\newblock In \emph{2013 IEEE/ACM 6th International Conference on Utility and Cloud Computing}, pages 123--130. IEEE, 2013.

\bibitem[Wang et~al.(2023)Wang, Li, Dai, Chen, Zhou, Meng, Zhou, and Sun]{wang2023label}
Lean Wang, Lei Li, Damai Dai, Deli Chen, Hao Zhou, Fandong Meng, Jie Zhou, and Xu~Sun.
\newblock Label words are anchors: An information flow perspective for understanding in-context learning.
\newblock \emph{arXiv preprint arXiv:2305.14160}, 2023.

\bibitem[Wu et~al.(2024)Wu, Wang, Xiao, Peng, and Fu]{wu2024retrieval}
Wenhao Wu, Yizhong Wang, Guangxuan Xiao, Hao Peng, and Yao Fu.
\newblock Retrieval head mechanistically explains long-context factuality.
\newblock \emph{arXiv preprint arXiv:2404.15574}, 2024.

\bibitem[Xiao et~al.(2023)Xiao, Tian, Chen, Han, and Lewis]{xiao2023efficient}
Guangxuan Xiao, Yuandong Tian, Beidi Chen, Song Han, and Mike Lewis.
\newblock Efficient streaming language models with attention sinks.
\newblock \emph{arXiv preprint arXiv:2309.17453}, 2023.

\bibitem[Yang et~al.(2024)Yang, Han, Gao, Hu, Zhang, and Zhao]{yang2024pyramidinfer}
Dongjie Yang, XiaoDong Han, Yan Gao, Yao Hu, Shilin Zhang, and Hai Zhao.
\newblock Pyramidinfer: Pyramid kv cache compression for high-throughput llm inference.
\newblock \emph{arXiv preprint arXiv:2405.12532}, 2024.

\bibitem[Yang et~al.(2018)Yang, Qi, Zhang, Bengio, Cohen, Salakhutdinov, and Manning]{yang2018hotpotqa}
Zhilin Yang, Peng Qi, Saizheng Zhang, Yoshua Bengio, William Cohen, Ruslan Salakhutdinov, and Christopher~D Manning.
\newblock Hotpotqa: A dataset for diverse, explainable multi-hop question answering.
\newblock In \emph{Proceedings of the 2018 Conference on Empirical Methods in Natural Language Processing}, pages 2369--2380, 2018.

\bibitem[Zhang et~al.(2024)Zhang, Sheng, Zhou, Chen, Zheng, Cai, Song, Tian, R{\'e}, Barrett, et~al.]{zhang2024h2o}
Zhenyu Zhang, Ying Sheng, Tianyi Zhou, Tianlong Chen, Lianmin Zheng, Ruisi Cai, Zhao Song, Yuandong Tian, Christopher R{\'e}, Clark Barrett, et~al.
\newblock H2o: Heavy-hitter oracle for efficient generative inference of large language models.
\newblock \emph{Advances in Neural Information Processing Systems}, 36, 2024.

\bibitem[Zhong et~al.(2021)Zhong, Yin, Yu, Zaidi, Mutuma, Jha, Hassan, Celikyilmaz, Liu, Qiu, et~al.]{zhong2021qmsum}
Ming Zhong, Da~Yin, Tao Yu, Ahmad Zaidi, Mutethia Mutuma, Rahul Jha, Ahmed Hassan, Asli Celikyilmaz, Yang Liu, Xipeng Qiu, et~al.
\newblock Qmsum: A new benchmark for query-based multi-domain meeting summarization.
\newblock In \emph{Proceedings of the 2021 Conference of the North American Chapter of the Association for Computational Linguistics: Human Language Technologies}, pages 5905--5921, 2021.

\bibitem[Zhu et~al.(2023)Zhu, Yang, Wang, Song, Wu, Wei, and Li]{zhu2023pose}
Dawei Zhu, Nan Yang, Liang Wang, Yifan Song, Wenhao Wu, Furu Wei, and Sujian Li.
\newblock Pose: Efficient context window extension of llms via positional skip-wise training.
\newblock \emph{arXiv preprint arXiv:2309.10400}, 2023.

\end{thebibliography}

\appendix


\clearpage
\newpage

\section{Limitations}
\label{section:limitations}
Our experiments were limited to three base models: LLAMA-3-8B-Instruct, LLAMA-3-70B-Instruct and Mistral-7B-Instruct. While these models demonstrated consistent trends, the robustness of our findings could be enhanced by testing a broader array of model families, should resources permit. Additionally, our research was conducted exclusively in English, with no investigations into how these findings might be transferred to other languages. Expanding the linguistic scope of our experiments could provide a more comprehensive understanding of the applicability of our results globally.
Based on our results at LongBench and Needle-in-a-HayStack experiment, PyramidKV generally works decently in most of the language tasks (i.e., Single-Document QA, Multi-Document QA, Summerization, Few-Shot In-Context Learning, etc.). Although we observe that PyramidKV performs better in some tasks (i.e., Few-Shot In-Context Learning) compared with some other tasks (i.e., Summerization), we have not observed cases that the decoding result collapses at some tasks. This remains a new topic for future work to explore. 

\section{Future Work}
\label{section:future_work}
Our investigation on PyramidKV highlights considerable opportunities for optimizing KV cache compression by adjusting the number of KV caches retained according to the distinct attention patterns of each layer (or even for each head). For instance, the retention of KV cache for each layer could be dynamically modified based on real-time analysis of the attention matrices, ensuring that the compression strategy is consistently aligned with the changing attention dynamics within LLMs. Furthermore, our experiments indicate that PyramidKV significantly surpasses other methods in few-shot learning tasks, suggesting promising applications of KV cache in in-context learning. This approach could potentially enable the use of more shots within constrained memory limits.

\begin{figure}[ht]
    \centering
    \includegraphics[width=\columnwidth]{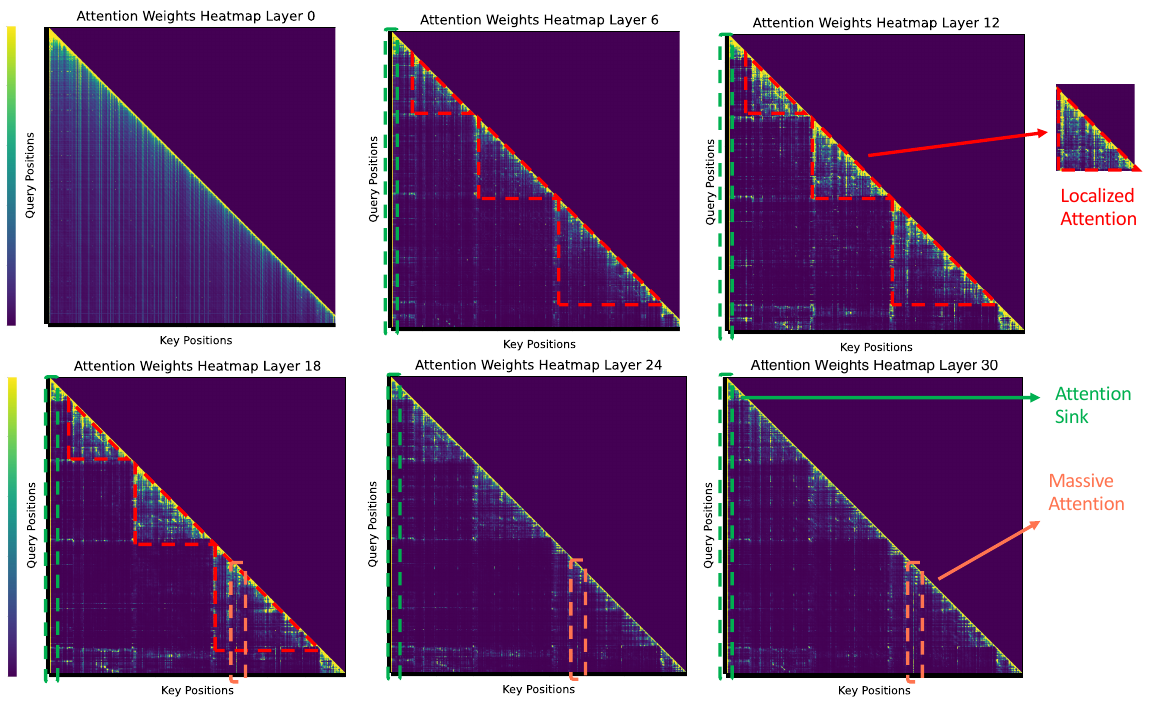}
    \caption{
  Attention patterns of retrieval-augmented generation across layers in 
Mistral-7B-Instruct model~\citep{jiang2023mistral}
}
    
    \label{figure:mistral_attention}
\end{figure}

\begin{figure}[ht]
    \centering
    \includegraphics[width=\columnwidth]{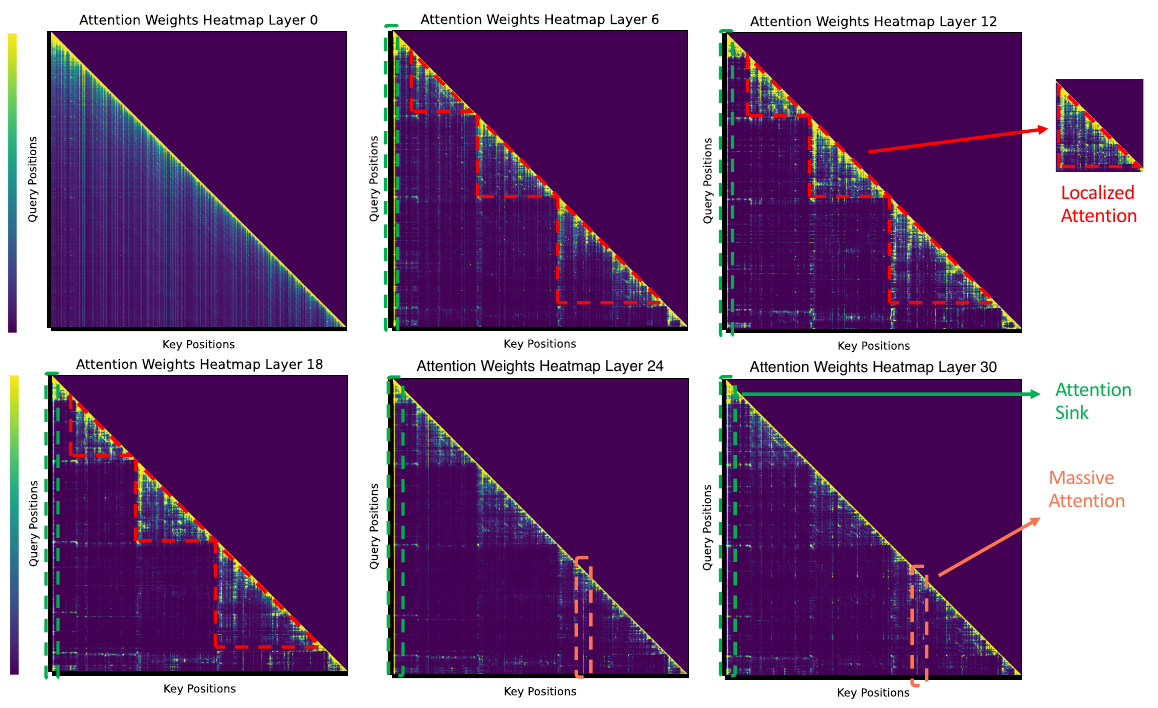}

  \caption{Attention patterns of retrieval-augmented generation across layers in Mixtral-8x7B-Instruct Mixture-of-Experts model.}

    \label{figure:mixtral_attention}
\end{figure}

\section{Related Work}
\label{appendix:related_work}

\paragraph{Interpretation of LLMs}
Prior research has shown that attention matrices in LLMs are typically sparse~\citep{chen2024image,xiao2023efficient,zhang2024h2o}, focusing disproportionately on a few tokens. For instance, \citet{xiao2023efficient} identified an ``attention sink'' phenomenon, where maintaining the Key and Value (KV) states of the first few tokens can substantially restore the performance of windowed attention, despite these tokens not being semantically crucial. Similarly, \citet{sun2024massive} identified a ``massive activations'' pattern, where a minority of activations show significantly larger values than others within LLMs. Interestingly, these values remain relatively constant across different inputs and act as critical bias terms in the model. 

Further explorations in this field reveal distinct patterns across various attention heads and layers. \citet{li2024snapkv} observed that certain attention heads consistently target specific prompt attention features during decoding. Additionally, \citet{wang2023label} discovered that in In-Context Learning scenarios, label words in demonstration examples serve as semantic anchors. In the lower layers of an 
LLM, shallow semantic information coalesces around these label words, which subsequently guide the LLMs’ final output predictions by serving as reference points. Recently,
\citet{wu2024retrieval} revealed that a special type of attention head, the so-called retrieval head, is largely responsible for retrieving information. Inspired by these findings that the attention mechanism exhibits varying behaviors across different layers, we discovered that ``Massive Activation'' does not consistently manifest across all layers in long context sequences; instead, it predominantly occurs in the upper layers. Additionally, we identified a novel trend of information aggregation specific to long-context inputs, which will be further explained in \Scref{section:observation}.



\paragraph{KV Cache Compression}
There has been a growing interest in addressing LLMs' memory constraints on processing long context inputs. FastGen~\citep{ge2023model} introduces an adaptive KV cache management strategy that optimizes memory use by tailoring retention tactics to the specific nature of attention heads. This method involves evicting long-range contexts from heads that prioritize local interactions, discarding non-special tokens from heads focused on special tokens, and maintaining a standard KV cache for heads that engage broadly across tokens.  SnapKV~\citep{li2024snapkv} improves efficiency by compressing KV caches via selecting/clustering significant KV positions based on their attention scores.  Heavy Hitter Oracle (H2O)~\citep{zhang2024h2o} implements a dynamic eviction policy that effectively balances the retention of recent and historically significant tokens, optimizing memory usage while preserving essential information.  StreamingLLM~\citep{xiao2023efficient} enables LLMs trained on finite attention windows to handle infinite sequence lengths without fine-tuning, thus expanding the models' applicability to broader contexts. LM-Infinite~\citep{han2023lm} allows LLMs pre-trained with 2K or 4K-long segments to generalize to up to 200M length inputs while retaining perplexity without parameter updates.

While these approaches have significantly advanced the efficient management of memory for LLMs, they generally apply a fixed KV cache size across all layers. In contrast, our investigations into the attention mechanisms across different layers of LLMs reveal that the attention patterns vary from layer to layer, making a one-size-fits-all approach to KV cache management suboptimal. In response to this inefficiency, we propose a novel KV cache compression method, called \method that allocates different KV cache budgets across different layers, tailored to the unique demands and operational logic of each layer's attention mechanism. This layer-specific strategy takes a significant step toward balancing both memory efficiency and model performance, addressing a key limitation in existing methodologies.

\section{Pyramidal Information Funneling}
\label{appendix:observation}

\autoref{figure:mistral_attention} and \autoref{figure:mixtral_attention} shows the attention patterns of one QA example over six different layers (i.e., 0, 6, 12, 18, 24, and 30) for Mistral-7B-Instruct model and Mixtral-8x7B-Instruct Mixture-of-Experts model. 
\autoref{figure:mistral_attention} and \autoref{figure:mixtral_attention} demonstrate that the Pyramidal Information Funneling phenomenon is also evident in both the Mistral model and Mixtral model . The results reveal that, akin to Llama-like models, Mistral exhibit a progressively narrowing attention focus across layers. This supports the universality of the Pyramidal Information Funneling phenomenon across diverse model families. We hope this addresses your concern and underscores the generalizability of our findings.

Our analysis uniquely examines attention metrics across all transformer layers, from 0 to 30, leading to the discovery of a key phenomenon we term Pyramidal Information Funneling.

~\citet{lee2024infinigenefficientgenerativeinference} conducted a limited investigation into attention patterns, focusing only on the lower layer (layer 0) and a single upper layer (layer 18). While ~\citet{lee2024infinigenefficientgenerativeinference} noted that attention becomes more skewed in upper layers, it did not provide a fine-grained observation of attention patterns across all layers. In contrast, our study reveals several novel findings:

\begin{itemize}
    \item \textbf{Localized Attention}: We observe that attention progressively narrows its focus, targeting specific components within the input sequence.
    \item \textbf{Massive Attention Mechanism}: In the upper layers, attention heavily concentrates on a small set of critical tokens. Notably, these tokens are not limited to the leading positions, as observed in ~\citet{lee2024infinigenefficientgenerativeinference}, but also appear at regular intervals across the sequence. The discrepancy arises from differences in input settings, with ~\citet{lee2024infinigenefficientgenerativeinference} identifying massive attention only at the initial tokens.
\end{itemize}

These insights motivated us to propose a token-selection method based on the highest attention scores in the upper layers, rather than solely relying on tokens from earlier positions.

To the best of our knowledge, ~\citet{chen2024magicpiglshsamplingefficient} has not analyzed attention patterns across transformer layers.

Therefore, although ~\citet{lee2024infinigenefficientgenerativeinference} and ~\citet{chen2024magicpiglshsamplingefficient} are considered contemporaneous with our work, making a comparison unnecessary, the perspective of our observation is considered novel compared with ~\citet{lee2024infinigenefficientgenerativeinference} and ~\citet{chen2024magicpiglshsamplingefficient}. Moreover, although ~\citet{lee2024infinigenefficientgenerativeinference} also observed attention patterns, the method we proposed based on our observations is significantly different from ~\citet{lee2024infinigenefficientgenerativeinference}, further highlighting the novelty of our work.

\section{Details of Proposed Method}
\label{appendix:method}

\begin{figure}
    \centering
    \includegraphics[width=\columnwidth]{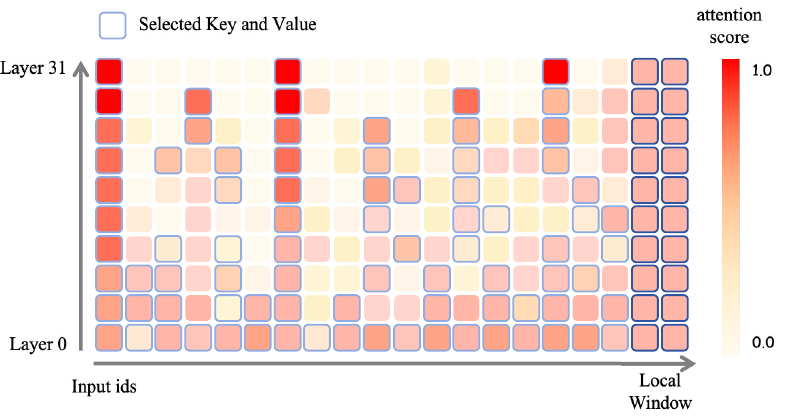}
    \caption{
    Illustration of~\method. At the lower level of the transformer, the \method selects more keys and values based on the exhibited average attention pattern. Fewer keys and values at the higher level are selected based on the massive activation pattern, where we observe that attention scores are concentrated over local regions.}
    \label{figure:method}
\end{figure}

Based on the pyramidal information funneling observed across different layers, \method consists of two steps: (1) Dynamically allocating different KV cache sizes/budgets across different layers; and (2) Selecting important KV vectors in each attention head for caching as ~\autoref{figure:method}.

Our decision to use an arithmetic sequence is driven by three key factors:
\begin{itemize}
    \item \textbf{Alignment with Pyramidal Information Funneling Pattern}: Empirical observations reveal a pyramidal information funneling pattern, where lower layers exhibit dispersed attention while higher layers concentrate on fewer tokens. Inspired by this, we adopt the arithmetic sequence design to align with this natural progression.
    \item \textbf{Superior Empirical Performance}: Through extensive experimentation across diverse datasets, we compared various methods, including the arithmetic sequence and adaptive approaches. Results consistently showed that the arithmetic sequence method outperformed others.
    \item \textbf{Computational Efficiency}: The arithmetic sequence method introduces minimal computational overhead compared to adaptive approaches, which require dynamically computing cache budgets across layers.
\end{itemize}

To perform KV cache eviction, we use torch.gather. Below, we outline the memory allocation and release process of torch.gather:
\begin{itemize}
    \item \textbf{Index Selection}: Identify the positions of the elements to extract from the input tensor.
    \item \textbf{Memory Location Calculation}: Compute the specific memory locations of the elements to be extracted using the strides of the input tensor across each dimension.
    \item \textbf{Output Tensor Creation}: Allocate memory to create a new output tensor and copy the selected elements to their corresponding positions in the output tensor.
    \item \textbf{Memory Management}: Since torch.gather is not an in-place operation, it creates a new tensor to store the results, while the memory of the original input tensor is released.
\end{itemize}

The speed-up offered by PyramidKV is complementary to that achieved through tensor parallelism and pipeline parallelism, as these approaches are not mutually exclusive. PyramidKV can be seamlessly integrated with both tensor parallelism and pipeline parallelism.

\section{Details of Evaluation}
\label{appendix:evaluation}

We use LongBench~\citep{bai2023longbench} to assess the performance of \method on tasks involving long-context inputs. LongBench is a meticulously designed benchmark suite that tests the capabilities of language models in handling extended documents and complex information sequences. This benchmark was created for multi-task evaluation of long context inputs.

We present the details of metrics, language and data for LongBench at \autoref{table:stat}. 

We run all the experiments on NVIDIA A100.

\begin{table}[htbp]
\centering  
\resizebox{\textwidth}{!}{
\begin{tabular}{llrccc}
\toprule
Dataset & Source & Avg len & Metric & Language & \#data \\
\midrule
\emph{Single-Document QA} \\
NarrativeQA & Literature, Film & 18,409 & F1 & English & 200 \\
Qasper & Science & 3,619 & F1 & English & 200 \\
MultiFieldQA-en & Multi-field & 4,559 & F1 & English & 150 \\
\midrule
\emph{Multi-Document QA} \\
HotpotQA & Wikipedia & 9,151 & F1 & English & 200 \\
2WikiMultihopQA & Wikipedia & 4,887 & F1 & English & 200 \\
MuSiQue & Wikipedia & 11,214 & F1 & English & 200 \\
\midrule
\emph{Summarization} \\
GovReport & Government report & 8,734 & Rouge-L & English & 200 \\
QMSum & Meeting & 10,614 & Rouge-L & English & 200 \\
MultiNews & News & 2,113 & Rouge-L & English & 200 \\
\midrule
\emph{Few-shot Learning} \\
TREC & Web question & 5,177 & Accuracy (CLS) & English & 200 \\
TriviaQA & Wikipedia, Web & 8,209 & F1 & English & 200 \\
SAMSum & Dialogue & 6,258 & Rouge-L & English & 200 \\
\midrule
\emph{Synthetic Task} \\
PassageCount & Wikipedia & 11,141 & Accuracy (EM) & English & 200 \\
PassageRetrieval-en & Wikipedia & 9,289 & Accuracy (EM) & English & 200 \\
\midrule
\emph{Code Completion} \\
LCC & Github & 1,235 & Edit Sim & Python/C\#/Java & 500 \\
RepoBench-P & Github repository & 4,206 & Edit Sim & Python/Java & 500 \\
\bottomrule
\end{tabular}
}
\caption{An overview of the dataset statistics in LongBench~\citep{bai2023longbench}. `Source' denotes the origin of the context. `Accuracy (CLS)' refers to classification accuracy, while `Accuracy (EM)' refers to exact match accuracy.}
\label{table:stat}
\end{table}

\section{License}
\label{appendix:license}
LongBench: MIT


\section{Handle Rotary Embedding after Tokens are Removed in~\method}

We keep the rotary embedding unchanged after tokens are removed, so that LLMs can still capture the exact position information even if the tokens are removed. StreamingLLM~\citep{xiao2023efficient} shows that rolling kv cache with the correct relative position is crucial for maintaining performance. This is because StreamingLLM is designed to mainly handle unlimited context sizes, where contexts exceed the LLM’s fixed context length. Without changing the rotary embedding after token removal, LLMs would receive rotary embedding of a non-monotonic position sequence. For example, after the first KV cache compression, LLMs might receive the input position embedding as [$0, 1, 2, 3, 3096, 3097, \cdots, 4096$], and the position embedding of the generated sequences could be [$1005, 1006, 1007, \cdots$]. The position sequence of [$0, 1, 2, 3, 3096, …, 4096, 1005, 1006, 1007, \cdots$] is a non-monotonic sequence, which may negatively hurts the performance. In contrast, our targeting settings will not process unlimited context size. For example, given a input sequence of $4012$ length, after KV cache compression, the position sequence would be [$0, 4, 6, 16, \cdots, 3927, 3987, 4012$], and the position sequence of the generated tokens would be [$4013, 4014, \cdots$]. By keeping the rotary embedding unchanged after the tokens are removed, the LLM avoids non-monotonic position sequences, and the LLM can capture the exact position information even if the tokens are shifted. Our preliminary results show that rolling KV cache with the correct relative position will slightly decrease the performance.

\section{Ablation Study}

In this section, we present an ablation study for hyperparameters and allocation strategies.

Based on our observations of the attention pattern, we find that a relatively stable, linear arithmetic decrease aligns more closely with the underlying structure of the pattern. We conduct experiments comparing various allocation strategies.

We conducted hyperparameter testing on the original development sets of 16 datasets in LongBench. The parameter $\beta$ demonstrated remarkable stability, showing minimal sensitivity to varying hyperparameter settings, which highlights its robustness. Conversely, $\alpha$ consistently produced superior results when set to 8 or 16. Consequently, these values were adopted for subsequent experiments.
In Appendix H.2 and H.3, we further analyzed the impact of hyperparameter selection on KV cache budget allocation across different layers. The experiments reaffirmed that $\beta$  had negligible influence on the outcomes, underscoring its stability. Meanwhile, $\alpha$ continued to deliver optimal results at values of 8 and 16.

\subsection{Allocation Srategies}

Based on our observations of the attention pattern, we find that a relatively stable, linear arithmetic decrease aligns more closely with the underlying structure of the pattern.

We conduct experiments comparing various pyramidal allocation strategies (i.e., linear decay strategy, geometric decay strategy and exponential decay strategy) with a cache size of 64 as \autoref{table:ablation_decay} to confirm that a linear strategy is indeed optimal or preferable.

We also propose three adaptive allocation baselines, which are based on the entropy, Gini coefficient, and sparsity of the attention values at each layer. The weight of each layer is calculated based on its corresponding metric (entropy, Gini coefficient, or sparsity), and the budget is allocated accordingly. Specifically:
\begin{itemize}
    \item \textbf{Entropy-based allocation}: Layers with higher entropy receive higher weights. Each layer's entropy is calculated based on the the layer's attention.
    \item \textbf{Gini coefficient-based allocation}: Layers with higher Gini coefficients receive higher weights. Each layer's Gini coefficient is calculated based on the the layer's attention
\end{itemize}

The empirical results as \autoref{table:ablation_decay} consistently showed that the linear strategy outperformed its counterparts, establishing it as the most effective approach for our use case.
The experiment strengthens the rationale for choosing the specific allocation method.

\begin{table*}[htbp]
\centering
\resizebox{\textwidth}{!}{
\begin{tabular}{l@{\hspace{0.05ex}}c@{\hspace{0.05ex}}c@{\hspace{0.05ex}}c@{\hspace{0.05ex}}c@{\hspace{0.05ex}}c@{\hspace{0.05ex}}c@{\hspace{0.05ex}}c@{\hspace{0.05ex}}c@{\hspace{0.05ex}}c@{\hspace{0.05ex}}c@{\hspace{0.05ex}}c@{\hspace{0.05ex}}c@{\hspace{0.05ex}}c@{\hspace{0.05ex}}c@{\hspace{0.6ex}}c@{\hspace{0.6ex}}c@{\hspace{0.6ex}}c}
\specialrule{1pt}{0pt}{2pt}
\multirow{4}{*}{Stra.}  & \multicolumn{3}{c}{Single-Document QA} & \multicolumn{3}{c}{Multi-Document QA}& \multicolumn{3}{c}{Summarization}& \multicolumn{3}{c}{Few-shot Learning}& \multicolumn{2}{c}{Synthetic} & \multicolumn{2}{c}{Code} & Avg. \\
\cmidrule(lr){2-4}\cmidrule(lr){5-7}\cmidrule(lr){8-10}\cmidrule(lr){11-13}\cmidrule(lr){14-15}\cmidrule(lr){16-17}
& \rotatebox[origin=c]{30}{NrtvQA} & \rotatebox[origin=c]{30}{Qasper} & \rotatebox[origin=c]{30}{MF-en} & \rotatebox[origin=c]{30}{HotpotQA} & \rotatebox[origin=c]{30}{2WikiMQA} & \rotatebox[origin=c]{30}{Musique} & \rotatebox[origin=c]{30}{GovReport} & \rotatebox[origin=c]{30}{QMSum} & \rotatebox[origin=c]{30}{MultiNews} & \rotatebox[origin=c]{30}{TREC} & \rotatebox[origin=c]{30}{TriviaQA} & \rotatebox[origin=c]{30}{SAMSum} & \rotatebox[origin=c]{30}{PCount} & \rotatebox[origin=c]{30}{PRe} & \rotatebox[origin=c]{30}{Lcc} & \rotatebox[origin=c]{30}{RB-P} & \\
\midrule
Geo.  & 20.51&15.04&29.4&34.93&26.41&16.6&18.32&21.68&18.81&52&87.51&36.15&5.18&69.17&53.11&44.91&34.36 \\
Exp.  & 20.58&14.82&28.74&34.34&26.24&16.11&18.41&21.63&18.75&52.00&87.94&36.26&5.19&69.17&54.34&43.21&34.23 \\
Lin.  &21.13&14.18&30.26&35.12&23.76&16.17&18.33&21.65&19.23&58.00&88.31&37.07&5.23&69.50&52.61&45.74&34.76 \\
Entropy.  & 18.12&14.12&27.22&33.21&21.16&15.16&17.76&19.87&17.09&51&87.31&34.29&5.09&68.91&50.12&42.98&32.71 \\
Gini.  & 17.92&14.61&28.21&32.67&19.98&15.98&16.20&19.29&18.21&51.00&86.21&34.97&5.11&65.51&51.98&43.37&32.58 \\
\bottomrule
\end{tabular}
}
\caption{
Ablation study of allocation strategies.}
\label{table:ablation_decay}
\end{table*}

\subsection{Hyper Parameter $\alpha$}

We  present the study of $\alpha$ for LlaMa-3-8B-Instruct in 128 KV cache size budget at~\autoref{table:ablation_step_a}.We find that a small alpha value (i.e., 8, 16) leads to better performance than a larger alpha value (i.e., 24, 32, 40, 48).

\begin{table*}[htbp]

\centering

\resizebox{\textwidth}{!}{
\begin{tabular}{l@{\hspace{0.05ex}}c@{\hspace{0.05ex}}c@{\hspace{0.05ex}}c@{\hspace{0.05ex}}c@{\hspace{0.05ex}}c@{\hspace{0.05ex}}c@{\hspace{0.05ex}}c@{\hspace{0.05ex}}c@{\hspace{0.05ex}}c@{\hspace{0.05ex}}c@{\hspace{0.05ex}}c@{\hspace{0.05ex}}c@{\hspace{0.05ex}}c@{\hspace{0.05ex}}c@{\hspace{0.6ex}}c@{\hspace{0.6ex}}c@{\hspace{0.6ex}}c}

\specialrule{1pt}{0pt}{2pt}
\multirow{4}{*}{$\alpha$}  & \multicolumn{3}{c}{Single-Document QA} & \multicolumn{3}{c}{Multi-Document QA}& \multicolumn{3}{c}{Summarization}& \multicolumn{3}{c}{Few-shot Learning}& \multicolumn{2}{c}{Synthetic} & \multicolumn{2}{c}{Code} & Avg. \\
\cmidrule(lr){2-4}\cmidrule(lr){5-7}\cmidrule(lr){8-10}\cmidrule(lr){11-13}\cmidrule(lr){14-15}\cmidrule(lr){16-17}
& \rotatebox[origin=c]{30}{NrtvQA} & \rotatebox[origin=c]{30}{Qasper} & \rotatebox[origin=c]{30}{MF-en} & \rotatebox[origin=c]{30}{HotpotQA} & \rotatebox[origin=c]{30}{2WikiMQA} & \rotatebox[origin=c]{30}{Musique} & \rotatebox[origin=c]{30}{GovReport} & \rotatebox[origin=c]{30}{QMSum} & \rotatebox[origin=c]{30}{MultiNews} & \rotatebox[origin=c]{30}{TREC} & \rotatebox[origin=c]{30}{TriviaQA} & \rotatebox[origin=c]{30}{SAMSum} & \rotatebox[origin=c]{30}{PCount} & \rotatebox[origin=c]{30}{PRe} & \rotatebox[origin=c]{30}{Lcc} & \rotatebox[origin=c]{30}{RB-P} & \\
\midrule

8  & 21.40 & 16.92 & 31.62 & 38.45 & 28.72 & 18.59 & 19.96 & 22.49 & 20.96 & 66.50 & 89.35 & 38.43 & 5.92  & 69.00 & 57.86 & 51.80 & 37.37 \\ 
16 & 23.37 & 16.21 & 33.93 & 38.24 & 27.28 & 20.57 & 19.71 & 21.93 & 20.86 & 60.00 & 88.75 & 38.34 & 5.48  & 69.12 & 57.84 & 53.42 & 37.19 \\ 
24 & 22.85 & 14.51 & 32.26 & 38.38 & 28.36 & 20.33 & 19.55 & 21.72 & 20.72 & 54.50 & 88.71 & 38.46 & 5.48  & 69.50 & 56.83 & 53.65 & 36.61 \\ 
32 & 23.01 & 14.54 & 31.68 & 38.86 & 29.90 & 19.16 & 19.20 & 21.83 & 20.52 & 49.50 & 87.01 & 38.01 & 5.75  & 69.50 & 57.02 & 54.54 & 36.25 \\ 
40 & 21.70 & 13.06 & 30.14 & 36.78 & 27.34 & 18.88 & 18.72 & 21.37 & 19.79 & 44.00 & 87.74 & 38.43 & 6.08  & 69.25 & 56.11 & 53.89 & 35.21 \\ 
48 & 21.51 & 12.30 & 29.77 & 39.04 & 26.76 & 17.97 & 18.65 & 21.20 & 20.29 & 44.50 & 87.73 & 38.44 & 5.51  & 69.25 & 56.73 & 53.88 & 35.22 \\

\bottomrule
\end{tabular}
}
\caption{Ablation on $\alpha$.}
\label{table:ablation_step_a}
\end{table*}

\subsection{Hyper Parameter $\beta$}

One topic we want to analyze for our ablation study is the selection of $\beta$, which can determine the staircase. The smaller $\beta$ is, the gentler the staircase is; the larger $\beta$ is, the steeper the staircase is. We want to investigate the effect of  $\beta$ step size on the final result. Results on 128 KV cache size and LlaMa-3-8B-Instruct are shown in \autoref{table:ablation_step_b}. The results at \autoref{table:ablation_step_b} show that using a relatively small value of $\beta$ yields better outcomes, and~\method is generally robust to the selection of $\beta$.

\begin{table*}[htbp]

\centering

\resizebox{\textwidth}{!}{
\begin{tabular}{l@{\hspace{0.05ex}}c@{\hspace{0.05ex}}c@{\hspace{0.05ex}}c@{\hspace{0.05ex}}c@{\hspace{0.05ex}}c@{\hspace{0.05ex}}c@{\hspace{0.05ex}}c@{\hspace{0.05ex}}c@{\hspace{0.05ex}}c@{\hspace{0.05ex}}c@{\hspace{0.05ex}}c@{\hspace{0.05ex}}c@{\hspace{0.05ex}}c@{\hspace{0.05ex}}c@{\hspace{0.6ex}}c@{\hspace{0.6ex}}c@{\hspace{0.6ex}}c}

\specialrule{1pt}{0pt}{2pt}
\multirow{4}{*}{$\beta$}  & \multicolumn{3}{c}{Single-Document QA} & \multicolumn{3}{c}{Multi-Document QA}& \multicolumn{3}{c}{Summarization}& \multicolumn{3}{c}{Few-shot Learning}& \multicolumn{2}{c}{Synthetic} & \multicolumn{2}{c}{Code} & Avg. \\
\cmidrule(lr){2-4}\cmidrule(lr){5-7}\cmidrule(lr){8-10}\cmidrule(lr){11-13}\cmidrule(lr){14-15}\cmidrule(lr){16-17}
& \rotatebox[origin=c]{30}{NrtvQA} & \rotatebox[origin=c]{30}{Qasper} & \rotatebox[origin=c]{30}{MF-en} & \rotatebox[origin=c]{30}{HotpotQA} & \rotatebox[origin=c]{30}{2WikiMQA} & \rotatebox[origin=c]{30}{Musique} & \rotatebox[origin=c]{30}{GovReport} & \rotatebox[origin=c]{30}{QMSum} & \rotatebox[origin=c]{30}{MultiNews} & \rotatebox[origin=c]{30}{TREC} & \rotatebox[origin=c]{30}{TriviaQA} & \rotatebox[origin=c]{30}{SAMSum} & \rotatebox[origin=c]{30}{PCount} & \rotatebox[origin=c]{30}{PRe} & \rotatebox[origin=c]{30}{Lcc} & \rotatebox[origin=c]{30}{RB-P} & \\
\midrule
20 & 21.40 & 16.92 & 33.79 & 39.73 & 28.72 & 18.59 & 19.86 & 22.48 & 20.95 & 66.50 & 89.35 & 38.39 & 5.92 & 69.00 & 56.49 & 47.95 & 37.25 \\
18 & 21.71 & 16.24 & 33.59 & 39.89 & 27.94 & 18.38 & 19.76 & 22.32 & 21.20 & 66.50 & 88.98 & 38.93 & 5.46 & 69.50 & 56.47 & 49.23 & 37.25 \\
16 & 21.74 & 14.86 & 33.64 & 39.18 & 28.17 & 18.77 & 19.57 & 22.25 & 21.48 & 66.50 & 89.69 & 38.87 & 5.82 & 69.50 & 57.02 & 50.11 & 37.32 \\
14 & 22.53 & 16.31 & 33.50 & 40.50 & 28.15 & 19.26 & 19.66 & 22.39 & 21.38 & 65.50 & 90.02 & 38.56 & 5.75 & 69.50 & 57.51 & 49.71 & 37.51 \\
\bottomrule
\end{tabular}
}
\caption{Ablation on $\beta$.}
\label{table:ablation_step_b}
\end{table*}




\section{Integation with MInference}
\label{appendix:minference}

We would like to clarify that PyramidKV and MInference~\cite{jiang2024minference} are complementary approaches addressing different aspects of KV cache optimization. Specifically:
\begin{itemize}
    \item MInference focuses on accelerating the generation of KV caches during the pre-filling stage of LLM inference.
    \item In contrast, PyramidKV targets efficient KV cache management during LLM decoding.
\end{itemize}
To evaluate their respective strengths, we compared PyramidKV and MInference on Longbench using a KV cache size of 128. The results demonstrated the superior performance of PyramidKV.

Furthermore, we demonstrate that MInference and PyramidKV can be seamlessly integrated to achieve highly efficient inference while maintaining performance comparable to full attention. The results of MInference combined with PyramidKV, evaluated on Longbench with a KV cache size of 128, as PyramidKV + MInference hybrid approach.

\begin{table*}[htbp]
\centering
\resizebox{\textwidth}{!}{
\begin{tabular}{l@{\hspace{0.05ex}}c@{\hspace{0.05ex}}c@{\hspace{0.05ex}}c@{\hspace{0.05ex}}c@{\hspace{0.05ex}}c@{\hspace{0.05ex}}c@{\hspace{0.05ex}}c@{\hspace{0.05ex}}c@{\hspace{0.05ex}}c@{\hspace{0.05ex}}c@{\hspace{0.05ex}}c@{\hspace{0.05ex}}c@{\hspace{0.05ex}}c@{\hspace{0.05ex}}c@{\hspace{0.6ex}}c@{\hspace{0.6ex}}c@{\hspace{0.6ex}}c}
\specialrule{1pt}{0pt}{2pt}
\multirow{4}{*}{Stra.}  & \multicolumn{3}{c}{Single-Document QA} & \multicolumn{3}{c}{Multi-Document QA}& \multicolumn{3}{c}{Summarization}& \multicolumn{3}{c}{Few-shot Learning}& \multicolumn{2}{c}{Synthetic} & \multicolumn{2}{c}{Code} & Avg. \\
\cmidrule(lr){2-4}\cmidrule(lr){5-7}\cmidrule(lr){8-10}\cmidrule(lr){11-13}\cmidrule(lr){14-15}\cmidrule(lr){16-17}
& \rotatebox[origin=c]{30}{NrtvQA} & \rotatebox[origin=c]{30}{Qasper} & \rotatebox[origin=c]{30}{MF-en} & \rotatebox[origin=c]{30}{HotpotQA} & \rotatebox[origin=c]{30}{2WikiMQA} & \rotatebox[origin=c]{30}{Musique} & \rotatebox[origin=c]{30}{GovReport} & \rotatebox[origin=c]{30}{QMSum} & \rotatebox[origin=c]{30}{MultiNews} & \rotatebox[origin=c]{30}{TREC} & \rotatebox[origin=c]{30}{TriviaQA} & \rotatebox[origin=c]{30}{SAMSum} & \rotatebox[origin=c]{30}{PCount} & \rotatebox[origin=c]{30}{PRe} & \rotatebox[origin=c]{30}{Lcc} & \rotatebox[origin=c]{30}{RB-P} & \\
\midrule
PyramidKV  & 23.99&20.61&38.28&43.23&31.62&20.94&21.27&22.69&22.83&71&90.48&39.86&5.83&69.25&56.94&50.16&39.31 \\
MInference  & 19.74&30.63&40.41&44.28&35.22&20.65&28.43&23.35&26.75&72.00&87.90&42.78&6.30&64.00&58.76&5.06&38.86
 \\
M. $+$ P.  & 20.04&31.74&39.98&43.10&35.21&21.60&27.41&23.06&26.76&73.00&88.03&43.36&6.28&64.00&58.57&45.42&40.47 \\
\bottomrule
\end{tabular}
}
\caption{
Comparison between PyramidKV, MInference and MInference-PyramidKV hybrid method.}
\label{table:minference}
\end{table*}

In summary, we demonstrate that PyramidKV outperforms MInference on Longbench. Furthermore, when integrated with MInference, PyramidKV enhances its performance even further.

\section{Comparison with PyramidInfer}
\label{appendix:pyramidinfer}

Our work differs from PyramidInfer in two key aspects:
\begin{itemize}
    \item \textbf{Decay Strategy}: While PyramidInfer~\cite{yang2024pyramidinfer} employs a geometric decay strategy, our method adopts an arithmetic decay strategy. We argue that the relatively stable and linear nature of arithmetic decay better aligns with the behavior of the attention mechanism. This strategy is derived from empirically observed attention patterns, aiming to closely match them. Notably, our approach also achieves superior results, as demonstrated in the experimental results presented in the table below.
    \item \textbf{Token Selection}: PyramidInfer discards tokens in earlier layers, preventing them from being reconsidered in later layers. In contrast, our method allows previously discarded tokens to be re-evaluated in higher layers, recognizing that these tokens may still hold relevance at different stages of the model's processing.
    \item \textbf{Pyramidal Information Funneling Pattern}: A key contribution of our work lies in identifying and leveraging the pyramidal information funneling phenomenon within attention mechanisms. Through in-depth analysis, we observe that attention tends to disperse in earlier layers and progressively concentrates on crucial tokens in higher layers. This insight forms the foundation of our arithmetic decay strategy, ensuring that our method aligns more naturally with these intrinsic patterns.
\end{itemize}
Despite some similarities between the two approaches, these differences lead to significantly distinct outcomes. As shown in ~\autoref{table:pyramidinfer}, our method consistently outperforms PyramidInfer, highlighting the effectiveness of our design choices.

\begin{table*}[htbp]
\centering
\resizebox{\textwidth}{!}{
\begin{tabular}{l@{\hspace{0.05ex}}c@{\hspace{0.05ex}}c@{\hspace{0.05ex}}c@{\hspace{0.05ex}}c@{\hspace{0.05ex}}c@{\hspace{0.05ex}}c@{\hspace{0.05ex}}c@{\hspace{0.05ex}}c@{\hspace{0.05ex}}c@{\hspace{0.05ex}}c@{\hspace{0.05ex}}c@{\hspace{0.05ex}}c@{\hspace{0.05ex}}c@{\hspace{0.05ex}}c@{\hspace{0.6ex}}c@{\hspace{0.6ex}}c@{\hspace{0.6ex}}c}
\specialrule{1pt}{0pt}{2pt}
\multirow{4}{*}{Stra.}  & \multicolumn{3}{c}{Single-Document QA} & \multicolumn{3}{c}{Multi-Document QA}& \multicolumn{3}{c}{Summarization}& \multicolumn{3}{c}{Few-shot Learning}& \multicolumn{2}{c}{Synthetic} & \multicolumn{2}{c}{Code} & Avg. \\
\cmidrule(lr){2-4}\cmidrule(lr){5-7}\cmidrule(lr){8-10}\cmidrule(lr){11-13}\cmidrule(lr){14-15}\cmidrule(lr){16-17}
& \rotatebox[origin=c]{30}{NrtvQA} & \rotatebox[origin=c]{30}{Qasper} & \rotatebox[origin=c]{30}{MF-en} & \rotatebox[origin=c]{30}{HotpotQA} & \rotatebox[origin=c]{30}{2WikiMQA} & \rotatebox[origin=c]{30}{Musique} & \rotatebox[origin=c]{30}{GovReport} & \rotatebox[origin=c]{30}{QMSum} & \rotatebox[origin=c]{30}{MultiNews} & \rotatebox[origin=c]{30}{TREC} & \rotatebox[origin=c]{30}{TriviaQA} & \rotatebox[origin=c]{30}{SAMSum} & \rotatebox[origin=c]{30}{PCount} & \rotatebox[origin=c]{30}{PRe} & \rotatebox[origin=c]{30}{Lcc} & \rotatebox[origin=c]{30}{RB-P} & \\
\midrule
Pyramidinfer  & 20.42&12.77&25.21&35.81&25.83&16.88&18.27&21.78&18.52&51.00&88.54&35.76&5.61&69.25&53.21&44.12&33.94
 \\
PyramidKV  &21.13&14.18&30.26&35.12&23.76&16.17&18.33&21.65&19.23&58.00&88.31&37.07&5.23&69.50&52.61&45.74&34.76  \\
\bottomrule
\end{tabular}
}
\caption{
Comparison between PyramidKV and Pyramidinfer.}
\label{table:pyramidinfer}
\end{table*}

\section{PyramidKV will cause minimal extra inference overhead.}
\label{appendix:inference_overhead}

The allocation strategy and score-based selection add minimal complexity in the inference phase compared to the computation required for next-token predictions as~\autoref{table:inference_overhead}. Each row shows the setting of using a specific “[Prompt length, Generation length]” combination.  We show the inference speed comparison between total inference time, time for allocation strategy and time for score-based selection on LlaMa-3-8B-Instruct. Each cell is the latency measured in seconds. Furthermore, our budget allocation can be calculated before inference, requiring only a one-time computation. Thus, PyramidKV will cause minimal extra inference overhead.

\begin{table}[h!]
\centering
\small
\begin{tabular}{ccccccccc}
\toprule
\textbf{Prompt Length} & \textbf{Generation Length} & \textbf{Inference Time} & \textbf{Allocation Time} & \textbf{Selection Time} \\
\midrule
512&512 & 18.26 & 0.0000003 & 0.0194 \\
512&1024 & 34.69 & 0.000002 & 0.0133 \\
512&2048 & 70.69 & 0.000003 & 0.013 \\
512&4096 & 138.62 & 0.000005 & 0.013 \\
1024&512 & 17.32 & 0.000002 & 0.0131 \\
1024&1024 & 34.67 & 0.000002 & 0.01288 \\
1024&2048 & 70.21 & 0.000005 & 0.01296 \\
1024&4096 & 138.61 & 0.000003 & 0.01297 \\
2048&512 & 17.48 & 0.000004 & 0.0128 \\
2048&1024 & 34.78 & 0.000006 & 0.0129 \\
2048&2048 & 69.50 & 0.000003 & 0.01297 \\
2048&4096 & 138.59 & 0.000003 & 0.013 \\
4096&512 & 17.58 & 0.000002 & 0.013 \\
4096&1024 & 34.93 & 0.000004 & 0.0129 \\
4096&2048 & 69.65 & 0.000002 & 0.013 \\
4096&4096 & 138.87 & 0.000002 & 0.013 \\
\bottomrule
\end{tabular}
\caption{Extra inference overhead of~\method }
\label{table:inference_overhead}
\end{table}

\section{Inference Speed Comparison}
\label{appendix:inference_speed_comparison}

\method does not require extra computation time for budget allocation at inference by design. We show the inference speed comparison between~\method and baselines on LlaMa-3-8B-Instruct as~\autoref{table:inference_speed_comparison}. Each row shows the setting of using a specific “[Prompt length, Generation length]” combination. Each cell is the latency measured in seconds.~\method does not sacrifice the speed.~\method provides performance improvement and memory saving while runs at a comparable speed compared with baselines (i.e. SnapKV~\citep{li2024snapkv}, StreamingLLM~\citep{xiao2023efficient} and H2O~\citep{zhang2024h2o}). That’s because the allocation strategy requires very limited additional complexity in the inference/generation phase compared with computation required for generation as~\autoref{appendix:inference_overhead}.

\begin{table}[h!]
\centering
\small
\begin{tabular}{ccccccc}
\toprule
\textbf{Prompt Length} & \textbf{Generation Length} & \textbf{H2O} & \textbf{SnapKV} & \textbf{StreamingLLM} & \textbf{PyramidKV} \\ 
\midrule
512&512  & 18.47 & 18.25 & 18.96 & 18.26 \\
512&1024 & 35.10 & 34.76 & 36.20 & 34.69 \\
512&2048 & 70.21 & 69.60 & 72.35 & 70.69 \\ 
512&4096 & 140.80 & 139.42 & 146.37 & 138.62 \\
1024&512 & 17.63 & 17.34 & 18.12 & 17.32 \\
1024&1024 & 35.16 & 34.61 & 36.17 & 34.67 \\
1024&2048 & 71.02 & 69.17 & 72.37 & 70.21 \\
1024&4096 & 140.51 & 138.83 & 146.09 & 138.61 \\ 
2048&512 & 17.64 & 19.54 & 18.22 & 17.48 \\ 
2048&1024 & 35.09 & 34.76 & 36.29 & 34.78 \\ 
2048&2048 & 70.84 & 69.56 & 72.46 & 69.50 \\ 
2048&4096 & 140.16 & 139.55 & 145.22 & 138.59 \\ 
4096&512 & 17.75 & 17.67 & 18.40 & 17.58 \\ 
4096&1024 & 35.20 & 35.08 & 36.46 & 34.93 \\ 
4096&2048 & 70.02 & 69.26 & 72.58 & 69.65 \\ 
4096&4096 & 139.87 & 138.57 & 144.98 & 138.87 \\ 
\bottomrule
\end{tabular}
\caption{Performance comparison across different configurations and methods.}
\label{table:inference_speed_comparison}
\end{table}

\section{\method Excels in all KV Cache Size Limitation}
\label{appendix:results}

The evaluation results from LongBench\citep{bai2023longbench} are shown in \autoref{table:longbench_appendix_llama3_8b}, \autoref{table:longbench_appendix_mistral}, and\autoref{table:longbench_appendix_llama3_70b}.
We report the results using LlaMa-3-8B-Instruct, LlaMa-3-70B-Instruct and Mistral-7B-Instruct\citep{jiang2023mistral} for different KV cache sizes.

Overall, \method consistently surpasses other method across a range of KV cache sizes and different backbone models, with its performance advantages becoming particularly pronounced in memory-constrained environments. Upon examining specific tasks, \method demonstrates a notably superior performance on the TREC task, a few-shot question answering challenge. This suggests that the model effectively aggregates information from the few-shot examples, highlighting the potential for further investigation into in-context learning tasks.

\begin{table*}

\centering


\resizebox{\textwidth}{!}{
\begin{tabular}{l@{\hspace{0.05ex}}c@{\hspace{0.05ex}}c@{\hspace{0.05ex}}c@{\hspace{0.05ex}}c@{\hspace{0.05ex}}c@{\hspace{0.05ex}}c@{\hspace{0.05ex}}c@{\hspace{0.05ex}}c@{\hspace{0.05ex}}c@{\hspace{0.05ex}}c@{\hspace{0.05ex}}c@{\hspace{0.05ex}}c@{\hspace{0.05ex}}c@{\hspace{0.05ex}}c@{\hspace{0.6ex}}c@{\hspace{0.6ex}}c@{\hspace{0.6ex}}c}

\specialrule{1pt}{0pt}{2pt}
\multirow{5}{*}{Method}  & \multicolumn{3}{c}{Single-Document QA} & \multicolumn{3}{c}{Multi-Document QA}& \multicolumn{3}{c}{Summarization}& \multicolumn{3}{c}{Few-shot Learning}& \multicolumn{2}{c}{Synthetic} & \multicolumn{2}{c}{Code} & \multirow{5}{*}{Avg.} \\
\cmidrule(lr){2-4}\cmidrule(lr){5-7}\cmidrule(lr){8-10}\cmidrule(lr){11-13}\cmidrule(lr){14-15}\cmidrule(lr){16-17}
& \rotatebox[origin=c]{30}{NrtvQA} & \rotatebox[origin=c]{30}{Qasper} & \rotatebox[origin=c]{30}{MF-en} & \rotatebox[origin=c]{30}{HotpotQA} & \rotatebox[origin=c]{30}{2WikiMQA} & \rotatebox[origin=c]{30}{Musique} & \rotatebox[origin=c]{30}{GovReport} & \rotatebox[origin=c]{30}{QMSum} & \rotatebox[origin=c]{30}{MultiNews} & \rotatebox[origin=c]{30}{TREC} & \rotatebox[origin=c]{30}{TriviaQA} & \rotatebox[origin=c]{30}{SAMSum} & \rotatebox[origin=c]{30}{PCount} & \rotatebox[origin=c]{30}{PRe} & \rotatebox[origin=c]{30}{Lcc} & \rotatebox[origin=c]{30}{RB-P} & \\
\cmidrule(lr){2-17}
&18409&3619&4559&9151&4887&11214&8734&10614&2113&5177&8209&6258&11141&9289&1235&4206& \\

\midrule
\multicolumn{18}{c}{LlaMa-3-8B-Instruct, KV Size = Full} \\
\arrayrulecolor{black!20}\midrule
FKV & 25.70 & 29.75 & 41.12 & 45.55 & 35.87 & 22.35 & 25.63 & 23.03 & 26.21 & 73.00 & 90.56 & 41.88 & 04.67 & 69.25 & 58.05 & 50.77 & 41.46 \\

\arrayrulecolor{black}\midrule
\multicolumn{18}{c}{LlaMa-3-8B-Instruct, KV Size = 64} \\
\arrayrulecolor{black!20}\midrule
SKV & 19.86 & 9.09 & 27.89 & \textbf{37.34} & 28.35 & \textbf{18.17} & 15.86 & 20.80 & 16.41 & 38.50 & 85.92 & 36.32 & 5.22 & 69.00 & 51.78 & 48.38 & 33.05 \\
H2O & 20.80 & 11.34 & 27.03 & 37.25 & 30.01 & 17.94 & 18.29 & 21.49 & 19.13 & 38.00 & 84.70 & \textbf{37.76} & \textbf{5.63} & 69.33 & \textbf{53.44} & \textbf{50.15} & 33.89 \\
SLM & 17.44 & 8.68 & 22.25 & 35.37 & \textbf{31.51} & 15.97 & 15.46 & 20.06 & 14.64 & 38.00 & 72.33 & 29.10 & 5.42 & \textbf{69.50} & 46.14 & 45.09 & 30.43 \\
Ours & \textbf{21.13} & \textbf{14.18} & \textbf{30.26} & 35.12 & 23.76 & 16.17 & \textbf{18.33} & \textbf{21.65} & \textbf{19.23} & \textbf{58.00} & \textbf{88.31} & 37.07 & 5.23 & \textbf{69.50} & 52.61 & 45.74 & \textbf{34.76} \\

\arrayrulecolor{black}\midrule
\multicolumn{18}{c}{LlaMa-3-8B-Instruct, KV Size = 96} \\
\arrayrulecolor{black!20}\midrule
SKV  & 20.45  & 10.34  & 31.84  & 37.85  & 28.65  & \textbf{18.52}  & 17.90  & 21.26  & 19.07  & 41.50  & 86.95  & 37.82  & 5.08  & 69.12  & 54.69  & \textbf{51.31}  & 34.51  \\
H2O  & 21.55  & 11.21  & 28.73  & 37.66  & 30.12  & 18.47  & 19.57  & 21.57  & 20.44  & 38.50  & \textbf{87.63}  & \textbf{38.47}  & 5.60  & 69.00  & 54.51  & 50.16  & 34.57  \\
SLM  & 18.67  & 8.43   & 24.98  & 38.35  & \textbf{30.59}  & 16.37  & 17.33  & 19.84  & 18.41  & 41.00  & 73.92  & 29.38  & 5.80  & \textbf{69.50}  & 47.15  & 45.61  & 31.58  \\
Ours & \textbf{21.67}  & \textbf{15.10}  & \textbf{33.50}  & \textbf{39.73}  & 26.48  & 17.47  & \textbf{19.64}  & \textbf{22.28}  & \textbf{20.49}  & \textbf{61.50}  & 87.38  & 38.18  & \textbf{6.00}  & 69.25  & \textbf{55.30}  & 46.78  & \textbf{36.29}  \\

\arrayrulecolor{black}\midrule
\multicolumn{18}{c}{LlaMa-3-8B-Instruct, KV Size = 128} \\
\arrayrulecolor{black!20}\midrule
SKV & 21.19 & 13.55 & 32.64 & 38.75 & 29.64 & \textbf{18.73} & 18.98 & 21.62 & 20.26 & 45.00 & 88.36 & 37.64 & 5.13 & 68.85 & 55.84 & \textbf{51.82} & 35.50 \\
H2O & \textbf{22.12} & 13.20 & 31.61 & 37.79 & \textbf{32.71} & 18.45 & \textbf{20.32} & 22.02 & \textbf{21.10} & 38.50 & 87.75 & \textbf{39.14} & 5.83 & \textbf{69.50} & 55.06 & 50.97 & 35.37 \\
SLM & 18.61 & 9.65 & 25.99 & 37.95 & 29.39 & 16.34 & 18.03 & 20.11 & 20.08 & 43.50 & 74.08 & 29.86 & 5.90 & \textbf{69.50} & 47.47 & 45.60 & 32.00 \\
Ours & 21.40 & \textbf{16.92} & \textbf{33.79} & \textbf{39.73} & 28.72 & 18.59 & 19.86 & \textbf{22.48} & 20.95 & \textbf{66.50} & \textbf{89.35} & 38.39 & \textbf{5.92} & 69.00 & \textbf{56.49} & 47.95 & \textbf{37.25} \\



\arrayrulecolor{black}\midrule
\multicolumn{18}{c}{LlaMa-3-8B-Instruct, KV Size = 2048} \\
\arrayrulecolor{black!20}\midrule

SKV & \textbf{25.86} & 29.55 & \textbf{41.10} & \textbf{44.99} & \textbf{35.80} & 21.81 & 25.98 & \textbf{23.40} & 26.46 & \textbf{73.50} & \textbf{90.56} & 41.66 & 5.17 & \textbf{69.25} & 56.65 & 49.94 & 41.35 \\
SLM & 21.71 & 25.78 & 38.13 & 40.12 & 32.01 & 16.86 & 23.14 & 22.64 & \textbf{26.48} & 70.00 & 83.22 & 31.75 & \textbf{5.74} & 68.50 & 53.50 & 45.58 & 37.82 \\
H2O & 25.56 & 26.85 & 39.54 & 44.30 & 32.92 & 21.09 & 24.68 & 23.01 & 26.16 & 53.00 & \textbf{90.56} & 41.84 & 4.91 & \textbf{69.25} & 56.40 & 49.68 & 39.35 \\
Ours & 25.40 & \textbf{29.71} & 40.25 & 44.76 & 35.32 & \textbf{21.98} & \textbf{26.83} & 23.30 & 26.19 & 73.00 & \textbf{90.56} & \textbf{42.14} & 5.22 & \textbf{69.25} & \textbf{58.76} & \textbf{51.18} & \textbf{41.49} \\

\arrayrulecolor{black}\bottomrule
\end{tabular}
}
\caption{Performance comparison of ~\method (Ours) with SnapKV (SKV), H2O, StreamingLLM (SLM) and FullKV (FKV) on LongBench for LlaMa-3-8B-Instruct. ~\method generally outperforms other KV Cache compression methods across various KV Cache sizes and LLMs. The performance strengths of ~\method are more evident in small KV Cache sizes. Bold text represents the best performance.
}
\label{table:longbench_appendix_llama3_8b}
\end{table*}

\begin{table*}

\centering


\resizebox{\textwidth}{!}{
\begin{tabular}{l@{\hspace{0.05ex}}c@{\hspace{0.05ex}}c@{\hspace{0.05ex}}c@{\hspace{0.05ex}}c@{\hspace{0.05ex}}c@{\hspace{0.05ex}}c@{\hspace{0.05ex}}c@{\hspace{0.05ex}}c@{\hspace{0.05ex}}c@{\hspace{0.05ex}}c@{\hspace{0.05ex}}c@{\hspace{0.05ex}}c@{\hspace{0.05ex}}c@{\hspace{0.05ex}}c@{\hspace{0.6ex}}c@{\hspace{0.6ex}}c@{\hspace{0.6ex}}c}

\specialrule{1pt}{0pt}{2pt}
\multirow{5}{*}{Method}  & \multicolumn{3}{c}{Single-Document QA} & \multicolumn{3}{c}{Multi-Document QA}& \multicolumn{3}{c}{Summarization}& \multicolumn{3}{c}{Few-shot Learning}& \multicolumn{2}{c}{Synthetic} & \multicolumn{2}{c}{Code} & \multirow{5}{*}{Avg.} \\
\cmidrule(lr){2-4}\cmidrule(lr){5-7}\cmidrule(lr){8-10}\cmidrule(lr){11-13}\cmidrule(lr){14-15}\cmidrule(lr){16-17}
& \rotatebox[origin=c]{30}{NrtvQA} & \rotatebox[origin=c]{30}{Qasper} & \rotatebox[origin=c]{30}{MF-en} & \rotatebox[origin=c]{30}{HotpotQA} & \rotatebox[origin=c]{30}{2WikiMQA} & \rotatebox[origin=c]{30}{Musique} & \rotatebox[origin=c]{30}{GovReport} & \rotatebox[origin=c]{30}{QMSum} & \rotatebox[origin=c]{30}{MultiNews} & \rotatebox[origin=c]{30}{TREC} & \rotatebox[origin=c]{30}{TriviaQA} & \rotatebox[origin=c]{30}{SAMSum} & \rotatebox[origin=c]{30}{PCount} & \rotatebox[origin=c]{30}{PRe} & \rotatebox[origin=c]{30}{Lcc} & \rotatebox[origin=c]{30}{RB-P} & \\
\cmidrule(lr){2-17}
&18409&3619&4559&9151&4887&11214&8734&10614&2113&5177&8209&6258&11141&9289&1235&4206& \\

\midrule
\multicolumn{18}{c}{Mistral-7B-Instruct, KV Size = Full} \\
\arrayrulecolor{black!20}\midrule
FKV & 26.90 & 33.07 & 49.20 & 43.02 & 27.33 & 18.78 & 32.91 & 24.21 & 26.99 & 71.00 & 86.23 & 42.65 & 2.75 & 86.98 & 56.96 & 54.52 & 42.71 \\

\arrayrulecolor{black}\midrule
\multicolumn{18}{c}{Mistral-7B-Instruct, KV Size = 64} \\
\arrayrulecolor{black!20}\midrule

SKV & 16.94&17.17&39.51&\textbf{36.87}&22.26&15.18&14.75&20.35&21.45&37.50&\textbf{84.16}&37.28&4.50&61.13&42.40&38.44&30.72\\
SLM & 15.01&13.84&28.74&30.97&\textbf{24.50}&13.42&13.25&19.46&19.17&35.50&76.91&29.61&4.67&27.33&38.71&35.29&25.60\\
H2O & 18.19&19.04&37.40&30.18&22.22&13.77&16.60&\textbf{21.52}&\textbf{21.98}&37.00&81.02&\textbf{38.62}&\textbf{5.00}&\textbf{66.03}&43.54&\textbf{40.46}&30.88\\
Ours & \textbf{20.91}&\textbf{20.21}&\textbf{39.94}&33.57&22.87&\textbf{15.70}&\textbf{17.31}&21.23&21.41&\textbf{54.00}&81.98&36.96&3.58&60.83&\textbf{44.52}&37.99&\textbf{32.19}\\

\arrayrulecolor{black}\midrule
\multicolumn{18}{c}{Mistral-7B-Instruct, KV Size = 96} \\
\arrayrulecolor{black!20}\midrule

SKV & 19.92&18.80&\textbf{43.29}&\textbf{39.66}&23.08&\textbf{15.94}&16.65&21.26&21.47&43.50&83.48&39.74&4.00&60.10&45.53&41.12&32.47\\
SLM & 15.15&15.48&31.44&30.03&\textbf{23.93}&12.73&16.76&19.15&19.19&41.50&75.31&28.71&5.00&28.48&38.92&36.05&26.37\\
H2O & 19.44&20.81&38.78&32.39&21.51&14.43&17.68&\textbf{22.40}&\textbf{21.99}&38.00&82.51&\textbf{39.94}&\textbf{6.06}&\textbf{77.48}&45.18&\textbf{42.43}&32.67\\
Ours & \textbf{20.35}&\textbf{21.87}&41.15&34.94&21.85&15.81&\textbf{18.21}&21.66&21.43&\textbf{65.00}&\textbf{83.60}&39.60&4.50&67.80&\textbf{45.83}&39.38&\textbf{34.08}\\

\arrayrulecolor{black}\midrule
\multicolumn{18}{c}{Mistral-7B-Instruct, KV Size = 128} \\
\arrayrulecolor{black!20}\midrule

SKV & 19.16 & 21.46 & 43.52 & \textbf{38.60} & \textbf{23.35} & \textbf{16.09} & 17.66 & 21.84 & 21.47 & 47.50 & \textbf{84.15} & 40.24 & 5.00 & 69.31 & \textbf{46.98} & \textbf{42.97} & 34.96 \\
SLM & 16.57 & 14.68 & 32.40 & 30.19 & 22.64 & 12.34 & 18.08 & 18.96 & 19.19 & 43.50 & 74.22 & 29.02 & 4.50 & 29.48 & 39.23 & 36.16 & 27.57 \\
H2O & 21.20 & 21.90 & 41.55 & 33.56 & 21.28 & 12.93 & \textbf{18.59} & \textbf{22.61} & \textbf{21.99} & 39.00 & 82.37 & \textbf{40.44} & \textbf{6.00} & \textbf{83.19} & 46.41 & 42.66 & 34.73 \\
Ours & \textbf{21.75} & \textbf{22.03} & \textbf{44.32} & 34.06 & 22.79 & 15.77 & 18.58 & 21.89 & 21.43 & \textbf{66.00} & 83.46 & 39.75 & 4.50 & 66.90 & 46.96 & 41.28 & \textbf{35.72} \\

\arrayrulecolor{black}\midrule
\multicolumn{18}{c}{Mistral-7B-Instruct, KV Size = 2048} \\
\arrayrulecolor{black!20}\midrule

SKV &\textbf{25.89} & \textbf{32.93} & 48.56 & \textbf{42.96} & 27.42 & 19.02 & 26.56 & \textbf{24.47} & 26.69 & 70.00 & 86.27 & 42.57 & \textbf{5.50} & \textbf{88.90} & 50.42 & 46.72 & 41.56 \\
SLM & 20.31 & 26.64 & 45.72 & 35.25 & 24.31 & 12.20 & \textbf{27.47} & 21.57 & 24.51 & 68.50 & 71.95 & 31.19 & 5.00 & 22.56 & 43.38 & 37.08 & 32.35 \\
H2O & 25.76 & 31.10 & \textbf{49.03} & 40.76 & 26.52 & 17.07 & 24.81 & 23.64 & 26.60 & 55.00 & \textbf{86.35} & 42.48 & \textbf{5.50} & 88.15 & 49.93 & 46.57 & 39.95 \\
Ours & 25.53 & 32.21 & 48.97 & 42.26 & \textbf{27.50} & \textbf{19.36} & 26.60 & 23.97 & \textbf{26.73} & \textbf{71.00} & 86.25 & \textbf{42.94} & 4.50 & 87.90 & \textbf{53.12} & \textbf{47.21} & \textbf{41.63} \\

\arrayrulecolor{black}\bottomrule
\end{tabular}
}
\caption{Performance comparison of ~\method (Ours) with SnapKV (SKV), H2O, StreamingLLM (SLM) and FullKV (FKV) on LongBench for Mistral-7B-Instruct. ~\method generally outperforms other KV Cache compression methods across various KV Cache sizes and LLMs. The performance strengths of ~\method are more evident in small KV Cache sizes. Bold text represents the best performance.
}
\label{table:longbench_appendix_mistral}
\end{table*}

\begin{table*}[!t]

\centering


\resizebox{\textwidth}{!}{
\begin{tabular}{l@{\hspace{0.05ex}}c@{\hspace{0.05ex}}c@{\hspace{0.05ex}}c@{\hspace{0.05ex}}c@{\hspace{0.05ex}}c@{\hspace{0.05ex}}c@{\hspace{0.05ex}}c@{\hspace{0.05ex}}c@{\hspace{0.05ex}}c@{\hspace{0.05ex}}c@{\hspace{0.05ex}}c@{\hspace{0.05ex}}c@{\hspace{0.05ex}}c@{\hspace{0.05ex}}c@{\hspace{0.6ex}}c@{\hspace{0.6ex}}c@{\hspace{0.6ex}}c}

\specialrule{1pt}{0pt}{2pt}
\multirow{5}{*}{Method}  & \multicolumn{3}{c}{Single-Document QA} & \multicolumn{3}{c}{Multi-Document QA}& \multicolumn{3}{c}{Summarization}& \multicolumn{3}{c}{Few-shot Learning}& \multicolumn{2}{c}{Synthetic} & \multicolumn{2}{c}{Code} & \multirow{5}{*}{Avg.} \\
\cmidrule(lr){2-4}\cmidrule(lr){5-7}\cmidrule(lr){8-10}\cmidrule(lr){11-13}\cmidrule(lr){14-15}\cmidrule(lr){16-17}
& \rotatebox[origin=c]{30}{NrtvQA} & \rotatebox[origin=c]{30}{Qasper} & \rotatebox[origin=c]{30}{MF-en} & \rotatebox[origin=c]{30}{HotpotQA} & \rotatebox[origin=c]{30}{2WikiMQA} & \rotatebox[origin=c]{30}{Musique} & \rotatebox[origin=c]{30}{GovReport} & \rotatebox[origin=c]{30}{QMSum} & \rotatebox[origin=c]{30}{MultiNews} & \rotatebox[origin=c]{30}{TREC} & \rotatebox[origin=c]{30}{TriviaQA} & \rotatebox[origin=c]{30}{SAMSum} & \rotatebox[origin=c]{30}{PCount} & \rotatebox[origin=c]{30}{PRe} & \rotatebox[origin=c]{30}{Lcc} & \rotatebox[origin=c]{30}{RB-P} & \\
\cmidrule(lr){2-17}
&18409&3619&4559&9151&4887&11214&8734&10614&2113&5177&8209&6258&11141&9289&1235&4206& \\

\midrule
\multicolumn{18}{c}{LlaMa-3-70B-Instruct, KV Size = Full} \\
\arrayrulecolor{black!20}\midrule
FKV & 27.75& 	46.48& 	49.45	& 52.04& 	54.9& 	30.42& 	32.37& 	22.27	& 27.58	& 73.5& 	92.46& 	45.73	& 12.5	& 72.5	& 40.96	& 63.91	& 46.55 \\

\arrayrulecolor{black}\midrule
\multicolumn{18}{c}{LlaMa-3-70B-Instruct, KV Size = 64} \\
\arrayrulecolor{black!20}\midrule

SKV & 23.92&31.09&36.54&46.66&50.40&25.30&18.05&21.11&19.79&41.50&\textbf{91.06}&40.26&12.00&\textbf{72.50}&43.33&57.62&39.45\\
SLM & 22.07&23.53&27.31&43.21&\textbf{51.66}&23.85&16.62&19.74&15.20&39.50&76.89&33.06&12.00&\textbf{72.50}&40.23&50.20&35.47\\
H2O & 25.45&34.64&33.23&\textbf{48.25}&50.30&24.88&20.03&21.50&21.39&42.00&90.36&\textbf{41.58}&12.00&71.50&43.83&\textbf{58.16}&39.94\\
Ours & \textbf{25.47}&\textbf{36.71}&\textbf{42.29}&47.08&46.21&\textbf{28.30}&\textbf{20.60}&\textbf{21.62}&\textbf{21.62}&\textbf{64.50}&89.61&41.28&\textbf{12.50}&\textbf{72.50}&\textbf{45.34}&56.50&\textbf{42.01}\\

\arrayrulecolor{black}\midrule
\multicolumn{18}{c}{LlaMa-3-70B-Instruct, KV Size = 96} \\
\arrayrulecolor{black!20}\midrule

SKV & \textbf{25.78}&35.71&42.13&\textbf{50.38}&\textbf{51.46}&26.68&19.61&21.40&21.98&48.50&\textbf{92.11}&41.21&12.00&72.00&44.85&59.05&41.55\\
SLM & 23.31&29.46&29.21&41.85&45.92&23.00&18.42&19.71&18.57&45.00&76.79&33.54&12.00&\textbf{72.50}&40.49&50.73&36.28\\
H2O & 25.30&35.13&35.54&47.39&50.61&26.20&20.87&\textbf{21.80}&\textbf{22.93}&41.00&90.47&\textbf{43.42}&12.00&72.00&43.84&\textbf{59.86}&40.52\\
Ours & 25.47&\textbf{37.61}&\textbf{44.00}&47.33&45.36&\textbf{27.91}&\textbf{21.05}&21.60&22.31&\textbf{66.00}&91.45&42.36&12.00&\textbf{72.50}&\textbf{45.12}&56.88&\textbf{42.43}\\

\arrayrulecolor{black}\midrule
\multicolumn{18}{c}{LlaMa-3-70B-Instruct, KV Size = 128} \\
\arrayrulecolor{black!20}\midrule

SKV & \textbf{26.22}&37.49&\textbf{45.70}&\textbf{50.86}&\textbf{52.82}&28.50&20.38&21.72&22.56&53.00&91.61&41.43&12.00&71.50&45.06&\textbf{60.50}&42.58 \\
SLM & 24.25&29.12&29.24&40.20&46.28&21.80&19.55&19.42&20.61&48.00&76.60&33.21&12.00&\textbf{72.50}&40.65&51.03&36.53 \\
H2O & 25.61&35.02&37.74&47.77&51.16&26.87&20.57&20.78&23.33&42.00&91.65&43.85&12.00&\textbf{72.50}&43.50&59.67&40.88 \\
Ours & 26.06&\textbf{40.35}&45.67&50.20&52.78&\textbf{29.36}&\textbf{22.31}&\textbf{22.02}&\textbf{23.69}&\textbf{71.00}&\textbf{92.27}&\textbf{44.33}&12.00&\textbf{72.50}&\textbf{45.90}&59.55&\textbf{44.37} \\



\arrayrulecolor{black}\midrule
\multicolumn{18}{c}{LlaMa-3-70B-Instruct, KV Size = 2048} \\
\arrayrulecolor{black!20}\midrule

SKV & 26.73&45.18&47.91&\textbf{52.00}&\textbf{55.24}&30.48&28.76&22.35&27.31&72.50&92.38&45.58&12.00&\textbf{72.50}&\textbf{41.52}&\textbf{69.27}&46.36\\ 
SLM & 26.69&41.01&35.97&46.55&52.98&25.71&27.81&20.81&27.16&69.00&91.55&44.02&12.00&72.00&41.44&68.73&43.96\\ 
H2O & \textbf{27.67}&\textbf{46.51}&\textbf{49.54}&51.49&53.85&29.97&28.57&\textbf{22.79}&\textbf{27.53}&59.00&\textbf{92.63}&\textbf{45.94}&12.00&72.50&41.39&63.90&45.33\\ 
Ours & 27.22&46.19&48.72&51.62&54.56&\textbf{31.11}&\textbf{29.76}&22.50&27.27&\textbf{73.50}&91.88&45.47&12.00&72.50&41.36&69.12&\textbf{46.55}\\

\arrayrulecolor{black}\bottomrule
\end{tabular}
}
\caption{Performance comparison of ~\method (Ours) with SnapKV (SKV), H2O, StreamingLLM (SLM) and FullKV (FKV) on LongBench for LlaMa-3-70B-Instruct. ~\method generally outperforms other KV Cache compression methods across various KV Cache sizes and LLMs. The performance strengths of ~\method are more evident in small KV Cache sizes. Bold text represents the best performance.
}
\label{table:longbench_appendix_llama3_70b}
\end{table*}

\begin{figure}[ht]
    \centering
    \includegraphics[width=\columnwidth]{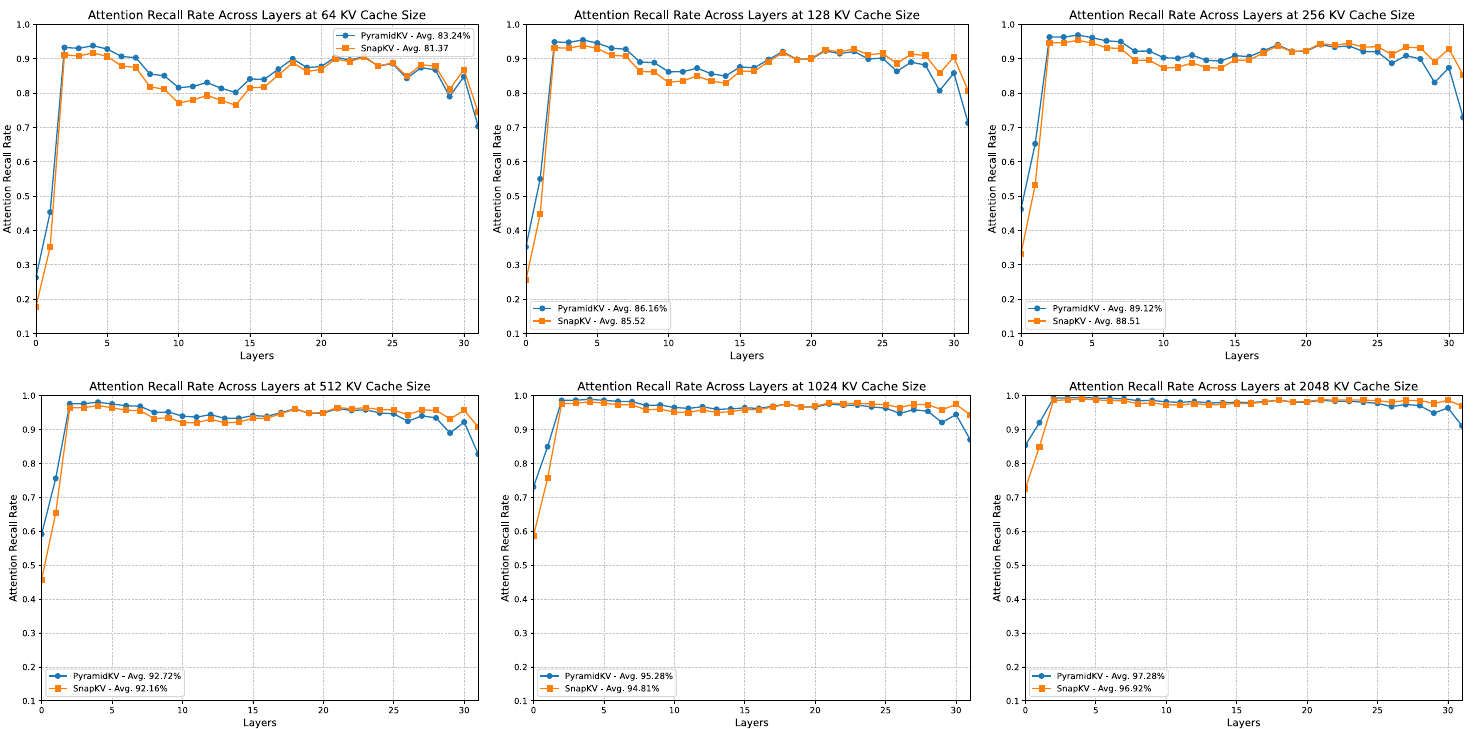}
    \caption{
  Attention recall rate (the percentage of attention computed using the keys retrieved by the method and the query, relative to the attention computed using all keys and the query.) comparison of PyramidKV and SnapKV.}
    \label{figure:attention_recall}
\end{figure}

With a small budget, our proposed method enables more effective allocation, better preserving useful attention information. Second, with a large budget, such allocation becomes less critical, as it is sufficient to cover the necessary information.
To further illustrate this phenomenon, we have included an ablation study titled "Attention Recall Rate Experiment" as ~\autoref{figure:attention_recall}. The results show that with a small budget, PyramidKV improves the attention recall rate (the percentage of attention computed using the keys retrieved by the method and the query, relative to the attention computed using all keys and the query.). However, with a larger budget (i.e., 2k KV Cache Size), the improvement decreases.
For 64, 128, 256, 512, 1024 and 2048 KV Cache sizes, PyramidKV's average attention recall rate improvements are 1.87\%, 0.64\%, 0.61\%, 0.56\%, 0.47\% and 0.36\%.

\section{LongBench results for 128 context length}
\label{appendix:128k}

We conducted additional experiments using Llama-3-8B-Instruct-Gradient-1048k with a sequence length of 128k as \autoref{table:128k}. The results, summarized in the table below, showcase the model's performance with extended context lengths. These findings provide further validation of the scalability and robustness of our approach.

\begin{table*}[htbp]
\centering
\resizebox{\textwidth}{!}{
\begin{tabular}{l@{\hspace{0.05ex}}c@{\hspace{0.05ex}}c@{\hspace{0.05ex}}c@{\hspace{0.05ex}}c@{\hspace{0.05ex}}c@{\hspace{0.05ex}}c@{\hspace{0.05ex}}c@{\hspace{0.05ex}}c@{\hspace{0.05ex}}c@{\hspace{0.05ex}}c@{\hspace{0.05ex}}c@{\hspace{0.05ex}}c@{\hspace{0.05ex}}c@{\hspace{0.05ex}}c@{\hspace{0.6ex}}c@{\hspace{0.6ex}}c@{\hspace{0.6ex}}c}
\specialrule{1pt}{0pt}{2pt}
\multirow{4}{*}{Method}  & \multicolumn{3}{c}{Single-Document QA} & \multicolumn{3}{c}{Multi-Document QA}& \multicolumn{3}{c}{Summarization}& \multicolumn{3}{c}{Few-shot Learning}& \multicolumn{2}{c}{Synthetic} & \multicolumn{2}{c}{Code} & Avg. \\
\cmidrule(lr){2-4}\cmidrule(lr){5-7}\cmidrule(lr){8-10}\cmidrule(lr){11-13}\cmidrule(lr){14-15}\cmidrule(lr){16-17}
& \rotatebox[origin=c]{30}{NrtvQA} & \rotatebox[origin=c]{30}{Qasper} & \rotatebox[origin=c]{30}{MF-en} & \rotatebox[origin=c]{30}{HotpotQA} & \rotatebox[origin=c]{30}{2WikiMQA} & \rotatebox[origin=c]{30}{Musique} & \rotatebox[origin=c]{30}{GovReport} & \rotatebox[origin=c]{30}{QMSum} & \rotatebox[origin=c]{30}{MultiNews} & \rotatebox[origin=c]{30}{TREC} & \rotatebox[origin=c]{30}{TriviaQA} & \rotatebox[origin=c]{30}{SAMSum} & \rotatebox[origin=c]{30}{PCount} & \rotatebox[origin=c]{30}{PRe} & \rotatebox[origin=c]{30}{Lcc} & \rotatebox[origin=c]{30}{RB-P} & \\
\midrule
SnapKV  & 
6.10&8.14&23.12&8.87&10.54&5.59&20.27&17.95&18.07&50.50&82.78&34.67&3.50&49.25&45.39&41.68&26.65
 \\
H2O  & 3.47&7.49&14.17&7.30&8.74&4.55&24.13&17.83&21.91&61.50&81.45&23.60&3.55&41.80&43.25&38.51&25.20
 \\
StreamingLLM  & 3.47&7.49&14.17&7.30&8.74&4.55&19.21&17.83&21.91&61.50&78.21&23.60&3.55&41.80&43.25&38.51&24.69 \\
PyramidKV  & 5.41&8.42&22.61&9.71&10.73&5.82&20.37&18.24&18.32&54.00&85.33&34.60&3.50&52.75&47.23&42.58&27.48 \\
\bottomrule
\end{tabular}
}
\caption{
Comparison of ~\method with baselines at 128k context length.}
\label{table:128k}
\end{table*}


\section{\method Preserves the Long-Context Understanding Ability}
\label{appendix:needle}

We perform Fact Retrieval Across Context Lengths (``Needle In A HayStack'')~\citep{liu2023lost,fu2024data} to test the in-context retrieval ability of LLMs after leveraging different KV cache methods. We conducted the Needle-in-a-Haystack experiment using various LLMs (i.e., Mistral-7B-Instruct-32k, LLaMA-3-8B-Instruct-8k, and LLaMA-3-70B-Instruct-8k), various KV cache sizes (i.e., 64, 96, and 128) and various methods (i.e., FullKV, PyramidKV, H2O and StreamingLLM). \method achieves Acc. performance closest to FullKV, while other methods show significant decreases. It is worth noting that PyramidKV with 128 KV cache size achieves the same 100.0 Acc. performance compared with FullKV with 8k context size for LLaMA-3-70B-Instruct.

\begin{figure}
    \centering
    \includegraphics[width=\columnwidth]{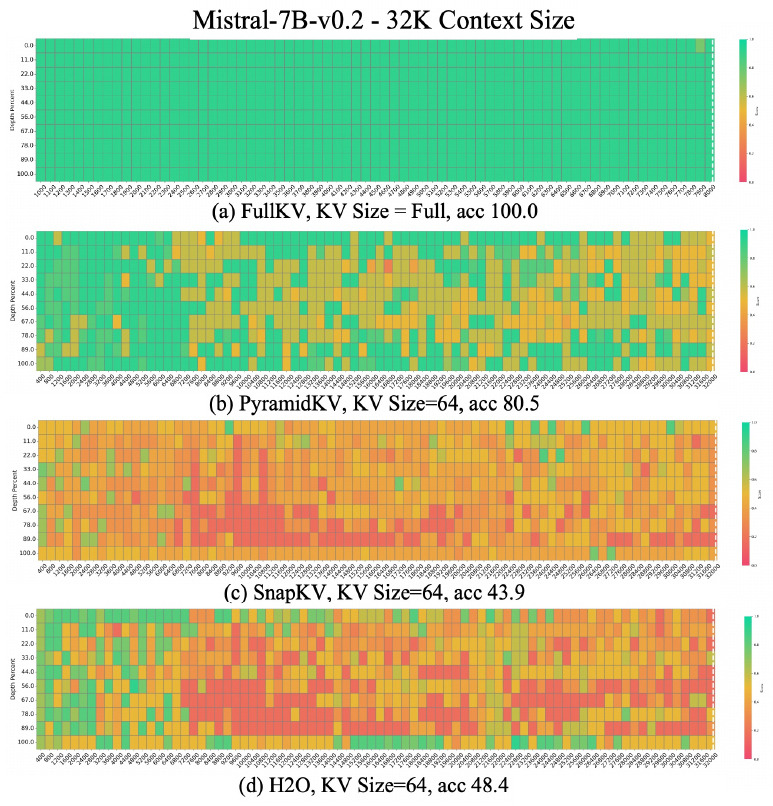}
    \caption{
    Results of the Fact Retrieval Across Context Lengths (``Needle In A HayStack'') test in \textbf{Mistral-7B-Instruct} with \textbf{32k} context size in \textbf{64} KV cache size.
    The vertical axis of the table represents the depth percentage, and the horizontal axis represents the token length. 
   ~\method mitigates the negative impact of KV cache compression on the long-context understanding capability of LLMs.
    }
    \label{figure:needle_mistral_64}
\end{figure}

\begin{figure}
    \centering
    \includegraphics[width=\columnwidth]{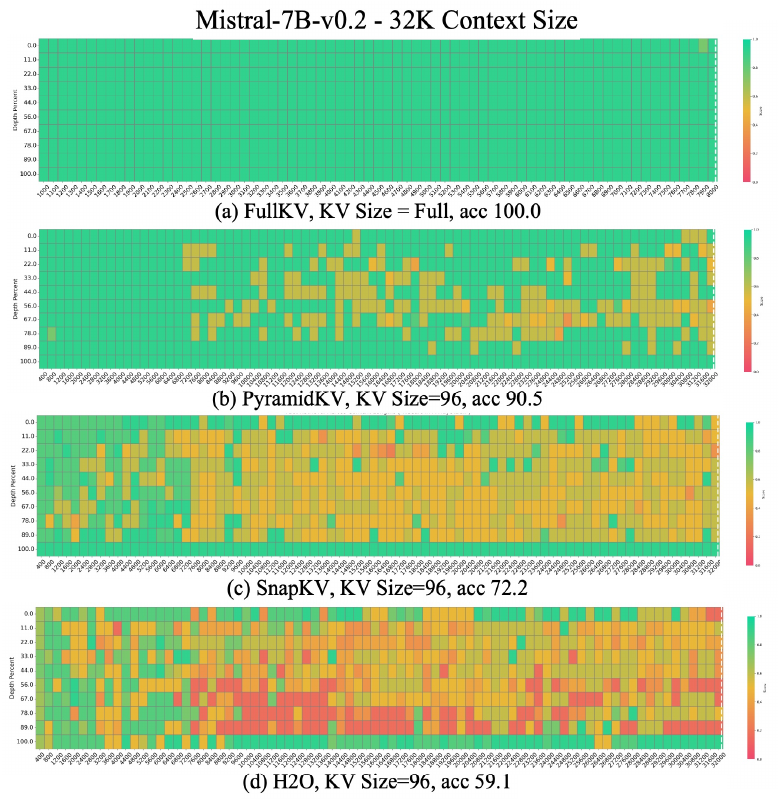}
    \caption{
    Results of the Fact Retrieval Across Context Lengths (``Needle In A HayStack'') test in \textbf{Mistral-7B-Instruct} with \textbf{32k} context size in \textbf{96} KV cache size.
    The vertical axis of the table represents the depth percentage, and the horizontal axis represents the token length. 
   ~\method mitigates the negative impact of KV cache compression on the long-context understanding capability of LLMs.
    }
    \label{figure:needle_mistral_96}
\end{figure}

\begin{figure}
    \centering
    \includegraphics[width=\columnwidth]{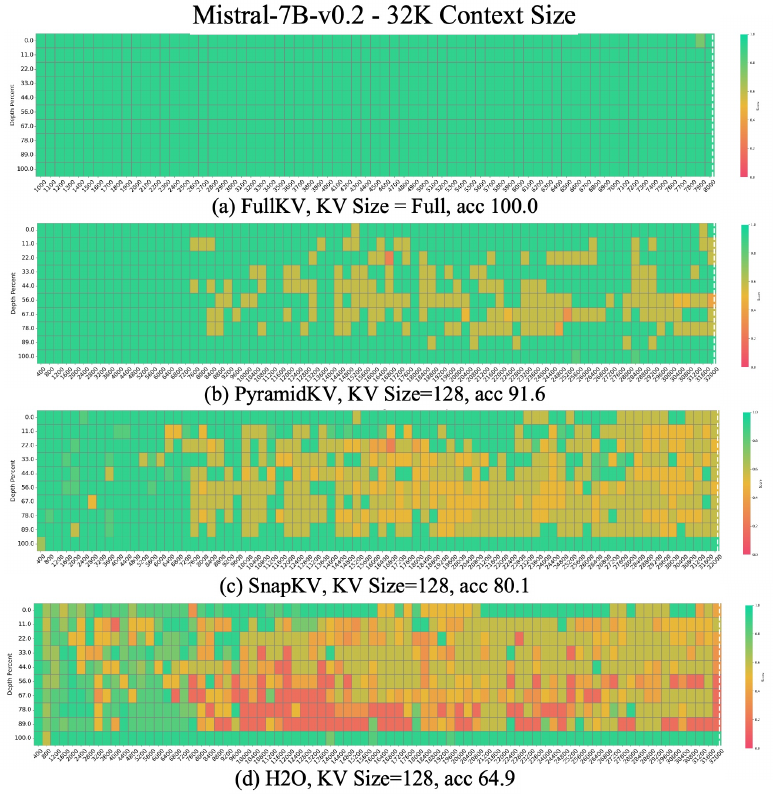}
    \caption{
    Results of the Fact Retrieval Across Context Lengths (``Needle In A HayStack'') test in \textbf{Mistral-7B-Instruct} with \textbf{32k} context size in \textbf{128} KV cache size.
    The vertical axis of the table represents the depth percentage, and the horizontal axis represents the token length. 
   ~\method mitigates the negative impact of KV cache compression on the long-context understanding capability of LLMs.
    }
    \label{figure:needle_mistral_128}
\end{figure}

\autoref{figure:needle_mistral_64}, \autoref{figure:needle_mistral_96}, \autoref{figure:needle_mistral_128} show the results of \textbf{Mistral-7B-Instruct}~\citep{jiang2023mistral} with different cache size (64, 96 and 128, respectively).

\begin{figure}
    \centering
    \includegraphics[width=\columnwidth]{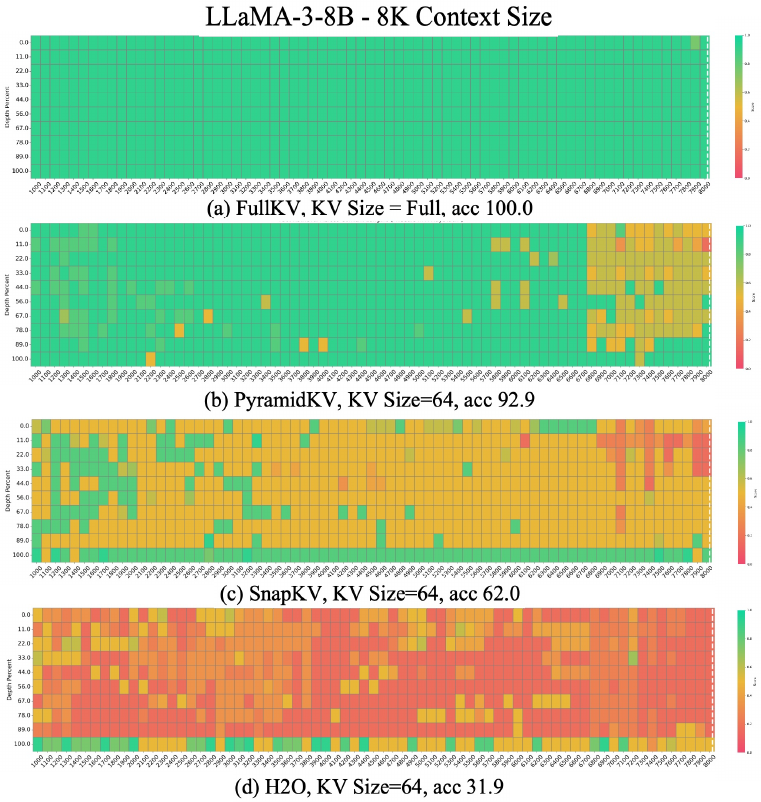}
    \caption{
    Results of the Fact Retrieval Across Context Lengths (``Needle In A HayStack'') test in \textbf{LlaMa-3-8B-Instruct} with \textbf{8k} context size in \textbf{64} KV cache size.
    The vertical axis of the table represents the depth percentage, and the horizontal axis represents the token length. 
   ~\method mitigates the negative impact of KV cache compression on the long-context understanding capability of LLMs.
    }
    \label{figure:needle_llama_64}
\end{figure}

\begin{figure}
    \centering
    \includegraphics[width=\columnwidth]{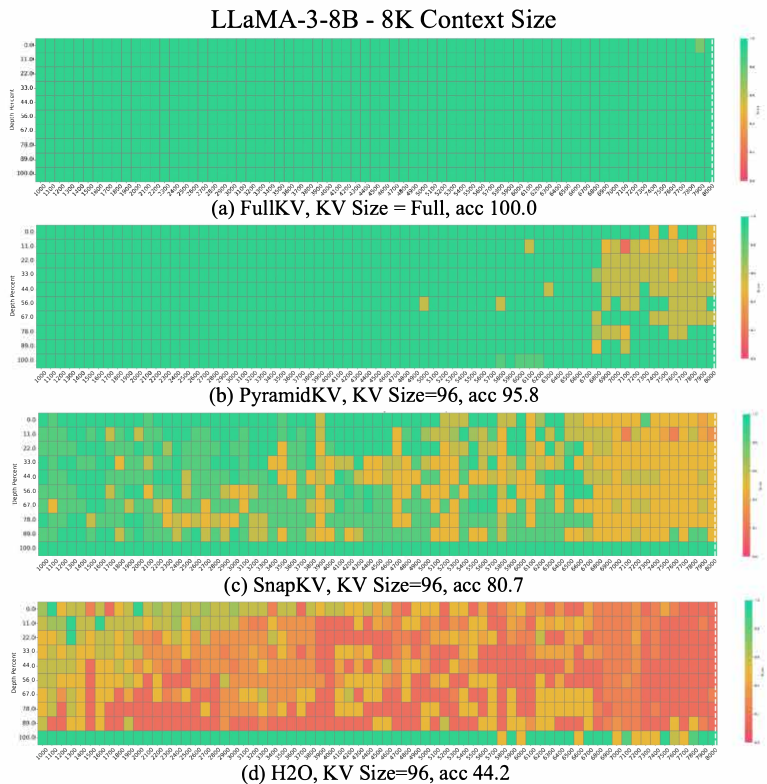}
    \caption{
    Results of the Fact Retrieval Across Context Lengths (``Needle In A HayStack'') test in \textbf{LlaMa-3-8B-Instruct} with \textbf{8k} context size in \textbf{96} KV cache size.
    The vertical axis of the table represents the depth percentage, and the horizontal axis represents the token length. 
   ~\method mitigates the negative impact of KV cache compression on the long-context understanding capability of LLMs.
    }
    \label{figure:needle_llama_96}
\end{figure}

\begin{figure}
    \centering
    \includegraphics[width=\columnwidth]{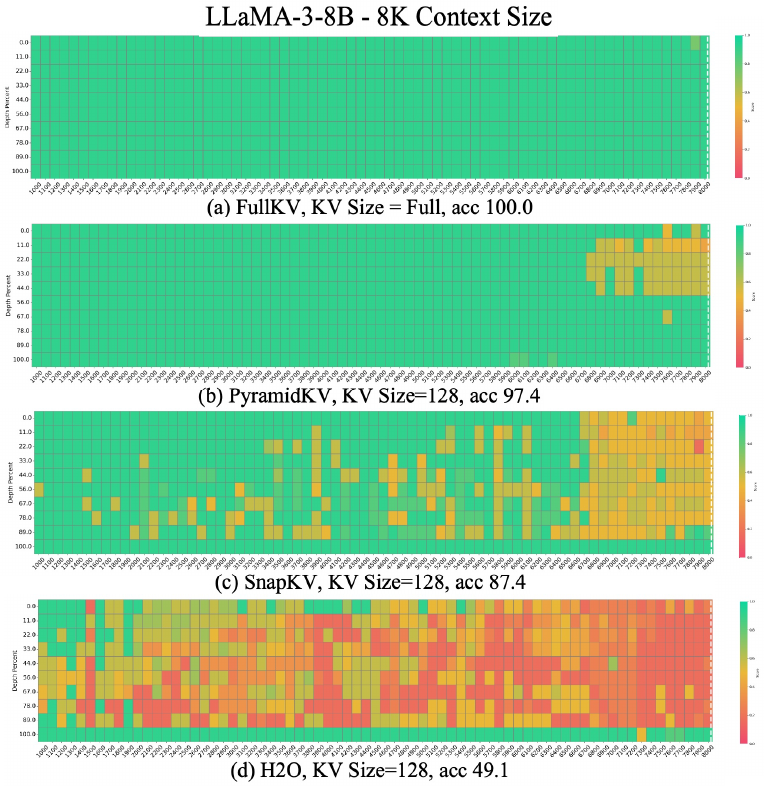}
    \caption{
    Results of the Fact Retrieval Across Context Lengths (``Needle In A HayStack'') test in \textbf{LlaMa-3-8B-Instruct} with \textbf{8k} context size in \textbf{128} KV cache size.
    The vertical axis of the table represents the depth percentage, and the horizontal axis represents the token length. 
   ~\method mitigates the negative impact of KV cache compression on the long-context understanding capability of LLMs.
    }
    \label{figure:needle_llama_128}
\end{figure}

\autoref{figure:needle_llama_64}, \autoref{figure:needle_llama_96}, \autoref{figure:needle_llama_128} show the results of \textbf{LlaMa-3-8B-Instruct} with different cache size (64, 96 and 128, respectively).

\begin{figure}
    \centering
    \includegraphics[width=\columnwidth]{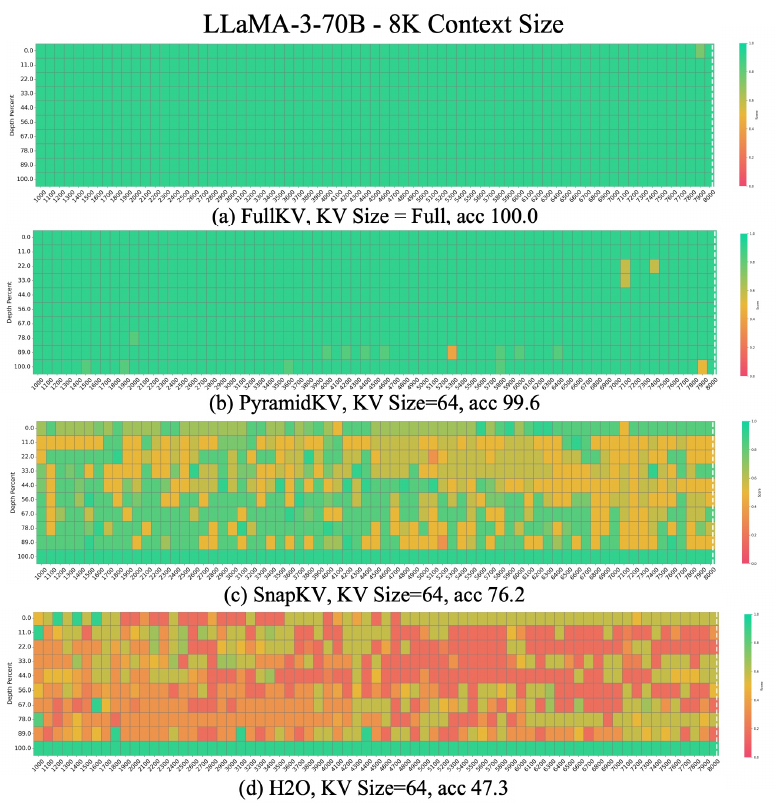}
    \caption{
    Results of the Fact Retrieval Across Context Lengths (``Needle In A HayStack'') test in \textbf{LlaMa-3-70B} with \textbf{8k} context size in \textbf{64} KV cache size.
    The vertical axis of the table represents the depth percentage, and the horizontal axis represents the token length. 
   ~\method mitigates the negative impact of KV cache compression on the long-context understanding capability of LLMs.
    }
    \label{figure:needle_llama70_64}
\end{figure}

\begin{figure}
    \centering
    \includegraphics[width=\columnwidth]{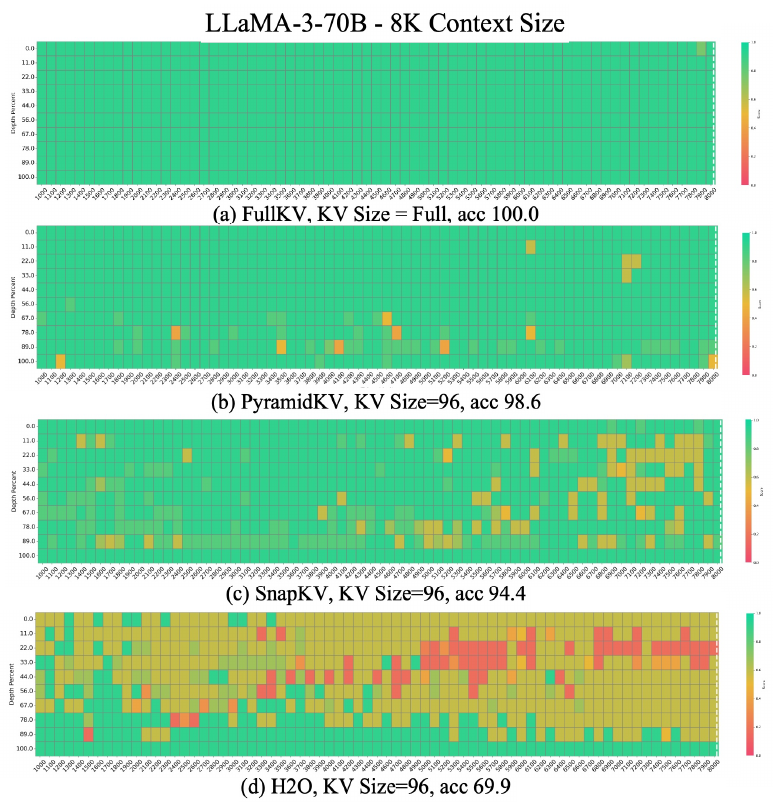}
    \caption{
    Results of the Fact Retrieval Across Context Lengths (``Needle In A HayStack'') test in \textbf{LlaMa-3-70B} with \textbf{8k} context size in \textbf{96} KV cache size.
    The vertical axis of the table represents the depth percentage, and the horizontal axis represents the token length. 
   ~\method mitigates the negative impact of KV cache compression on the long-context understanding capability of LLMs.
    }
    \label{figure:needle_llama70_96}
\end{figure}

\begin{figure}
    \centering
    \includegraphics[width=\columnwidth]{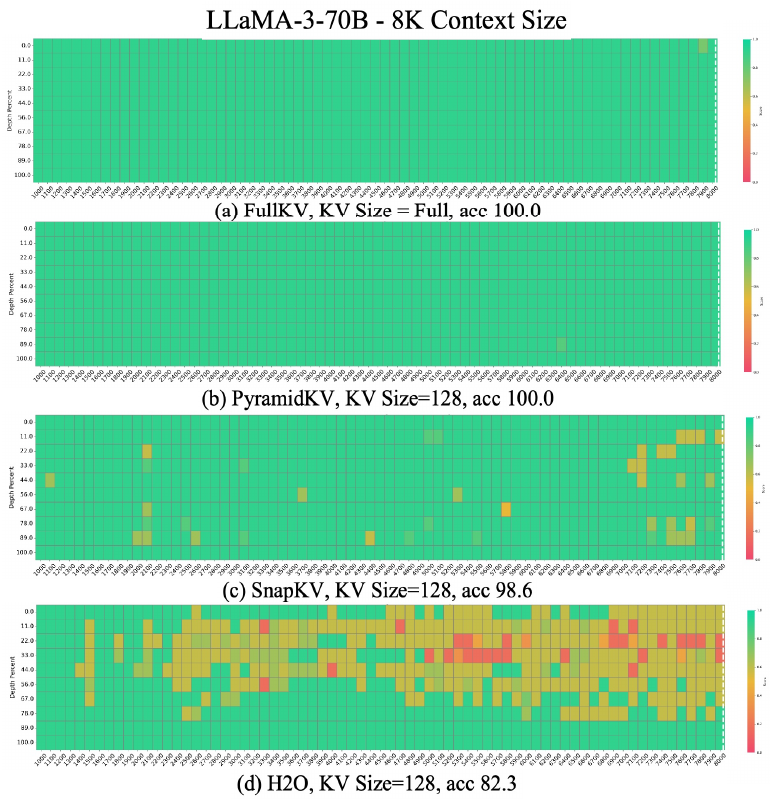}
    \caption{
    Results of the Fact Retrieval Across Context Lengths (``Needle In A HayStack'') test in \textbf{LlaMa-3-70B} with \textbf{8k} context size in \textbf{128} KV cache size.
    The vertical axis of the table represents the depth percentage, and the horizontal axis represents the token length. 
   ~\method mitigates the negative impact of KV cache compression on the long-context understanding capability of LLMs.
    }
    \label{figure:needle_llama70_128}
\end{figure}

\autoref{figure:needle_llama70_64}, \autoref{figure:needle_llama70_96}, \autoref{figure:needle_llama70_128} show the results of \textbf{LlaMa-3-70B-Instruct} with different cache size (64, 96 and 128, respectively).

\begin{table*}[h]
\small
\centering
\begin{tabular}{lcccccc}
\toprule
               Model & Length &  KV Cache &  Full KV Acc. &  PyramidKV Acc. &  SnapKV Acc. &  H2O Acc. \\
\midrule
 Mistral-7B &          32k &             64 &               100.00 &                  80.50 &               43.90 &            48.40 \\
 Mistral-7B &          32k &             96 &               100.00 &                  90.50 &               72.20 &            59.10 \\
 Mistral-7B &          32k &            128 &               100.00 &                  91.60 &               80.10 &            64.90 \\
 \midrule
 LLaMa-3-8B &           8k &             64 &               100.00 &                  92.90 &               62.00 &            31.90 \\
 LLaMa-3-8B &           8k &             96 &               100.00 &                  95.80 &               80.70 &            44.20 \\
 LLaMa-3-8B &           8k &            128 &               100.00 &                  97.40 &               87.40 &            49.10 \\
 \midrule
LLaMa-3-70B &           8k &             64 &               100.00 &                  99.60 &               76.20 &            47.30 \\
LLaMa-3-70B &           8k &             96 &               100.00 &                  98.60 &               94.40 &            69.90 \\
LLaMa-3-70B &           8k &            128 &               100.00 &                 100.00 &               98.60 &            82.30 \\
\bottomrule
\end{tabular}
\label{figure:needle}
\caption{Recall Accuracy performance from Fact Retrieval Across Context Lengths (``Needle In A HayStack'')
}
\vspace{-3mm}
\end{table*}

\clearpage
\newpage

\section{Attention Patterns across heads in the Bottom Layer}
\label{appendix:heads_attention}

\begin{figure}[ht]
    \centering
    \includegraphics[width=\columnwidth]{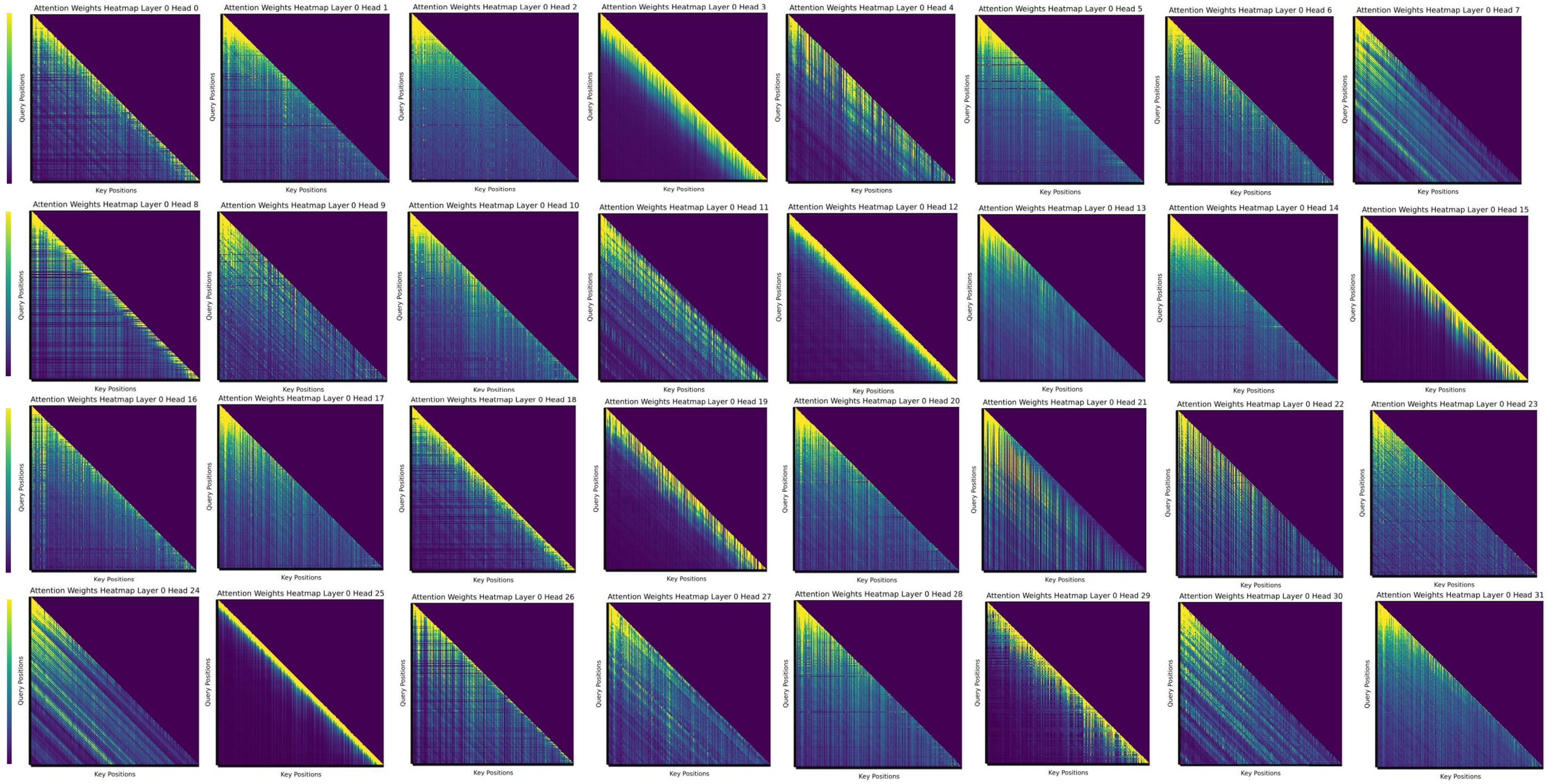}
    \caption{
  Attention patterns of retrieval-augmented generation across heads in the bottom layer in LlaMa.}
    \label{figure:attention_heads}
\end{figure}

Retrieval heads are predominantly located in the higher layers. Notably, no retrieval heads are observed in bottom layers.
To further investigate, we conducted additional experiments on the bottom layer to analyze the attention patterns of the heads as \autoref{figure:attention_heads}. Our findings indicate the absence of "massive attention" in any individual head.

\section{PyramidKV Implementation at vLLM}
\label{appendix:vllm}

To help compare the vLLM implementation with the vanilla dense attention backend in terms of throughput, we perform the experiment. We present the throughput comparison between the PyramidKV vLLM implementation and the vanilla dense attention backend in a setting where the inputs have varying context lengths without shared prefixes.

\begin{figure}[ht]
    \centering
    \includegraphics[width=\columnwidth]{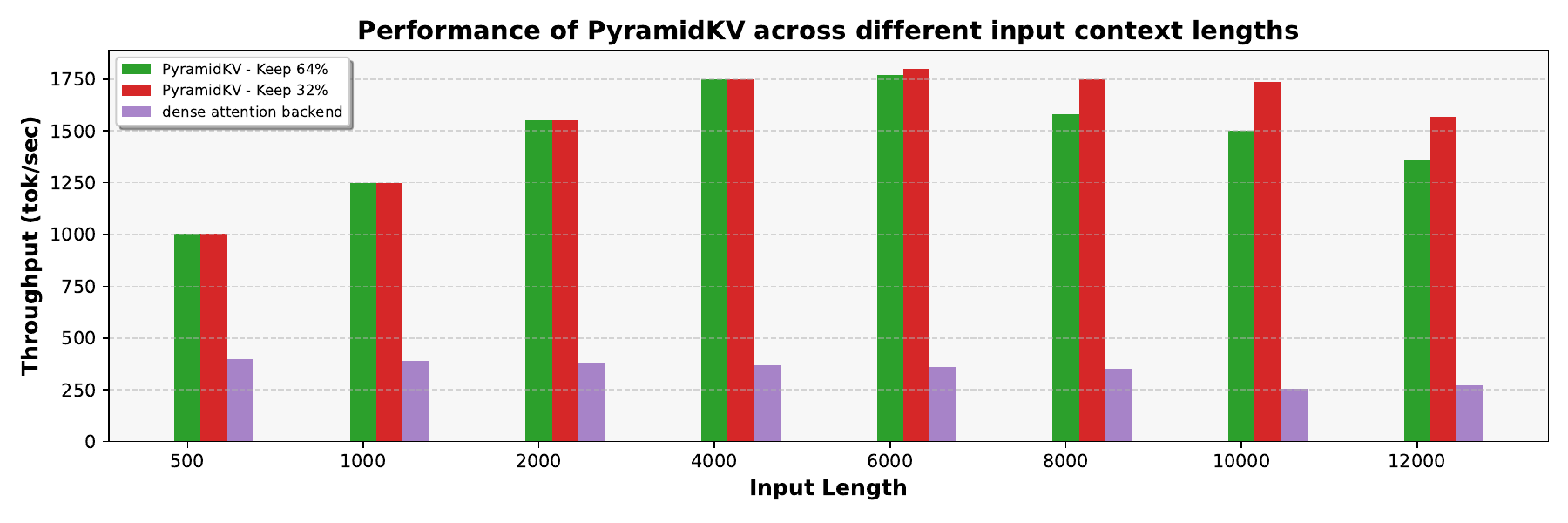}
    \caption{
    Throughout performance of PyramidKV across different input context lengths using LlaMa-3-8b model.
  }
    \label{figure:vllm_length}
\end{figure}

In Figure \autoref{figure:vllm_length}, we plot the throughput of the LlaMa 8b model by varying length. We observe that relative throughput under compression decreases as the new input context length approaches the limit, causing new sequences to wait longer before being added to the decoding batch.

We find that allocating/releasing/moving/accessing very small chunks of memory may cause inefficiency and fragmentation in a naive implementation of PyramidKV at vLLM. As PyramidKV applies different allocation budgets for different layers. The top layers have less budget, while the bottom layers have more budget. The application of KV cache eviction with different budgets across layers at the standard paged attention frameworks (i.e., vLLM) is ineffective as it only reduces the cache size proportionally to the layer with the lowest compression rate, and all evictions beyond this rate merely increase cache fragmentation.

However, the problem could be solved by adapting paged attention to page out cache on a per-layer basis. We expand the block tables of each sequence to include block tables for each layer of the cache so that they can be retrieved for each layer’s KV cache during attention without the use of fixed memory offsets.




\clearpage
\newpage

\end{document}